\documentclass{article}
\makeatletter
\def\input@path{{styles/}}
\makeatother
\usepackage{styles/fancyhdr,styles/natbib}
\usepackage[utf8]{inputenc}
\usepackage[T1]{fontenc}
\usepackage{CJKutf8}
\usepackage{styles/tongyi_report}

\usepackage{amsmath,amsfonts,bm}

\def\eqref#1{equation~\ref{#1}}

\def\1{\bm{1}}

\DeclareMathAlphabet{\mathsfit}{\encodingdefault}{\sfdefault}{m}{sl}
\SetMathAlphabet{\mathsfit}{bold}{\encodingdefault}{\sfdefault}{bx}{n}

\usepackage{booktabs}
\usepackage{colortbl}
\usepackage{graphicx}
\usepackage{wrapfig}
\usepackage{titletoc}
\usepackage{multicol}
\usepackage{capt-of}
\usepackage{makecell}
\usepackage{multirow}
\usepackage[section]{placeins}
\usepackage{xspace}
\usepackage{bm}
\usepackage{tikz}
\usepackage{pgfplots}
\pgfplotsset{compat=1.18}
\usetikzlibrary{arrows.meta,backgrounds,calc,fit,positioning,shapes.geometric}
\usepackage{hyperref}
\usepackage{url}
\usepackage{fontawesome5}

\definecolor{egoblue}{HTML}{0072B2}
\definecolor{egogreen}{HTML}{009E73}
\definecolor{egoorange}{HTML}{E69F00}
\definecolor{egopurple}{HTML}{7A5195}
\definecolor{referencepurple}{HTML}{513ECB}
\newcommand{\transferinline}[1]{%
  \begin{tikzpicture}[x=1pt,y=1pt,baseline=-3pt,line cap=round,line join=round]
    \node[inner sep=0pt] at (0,0) {\includegraphics[viewport=120 0 710 1010,clip,height=13pt]{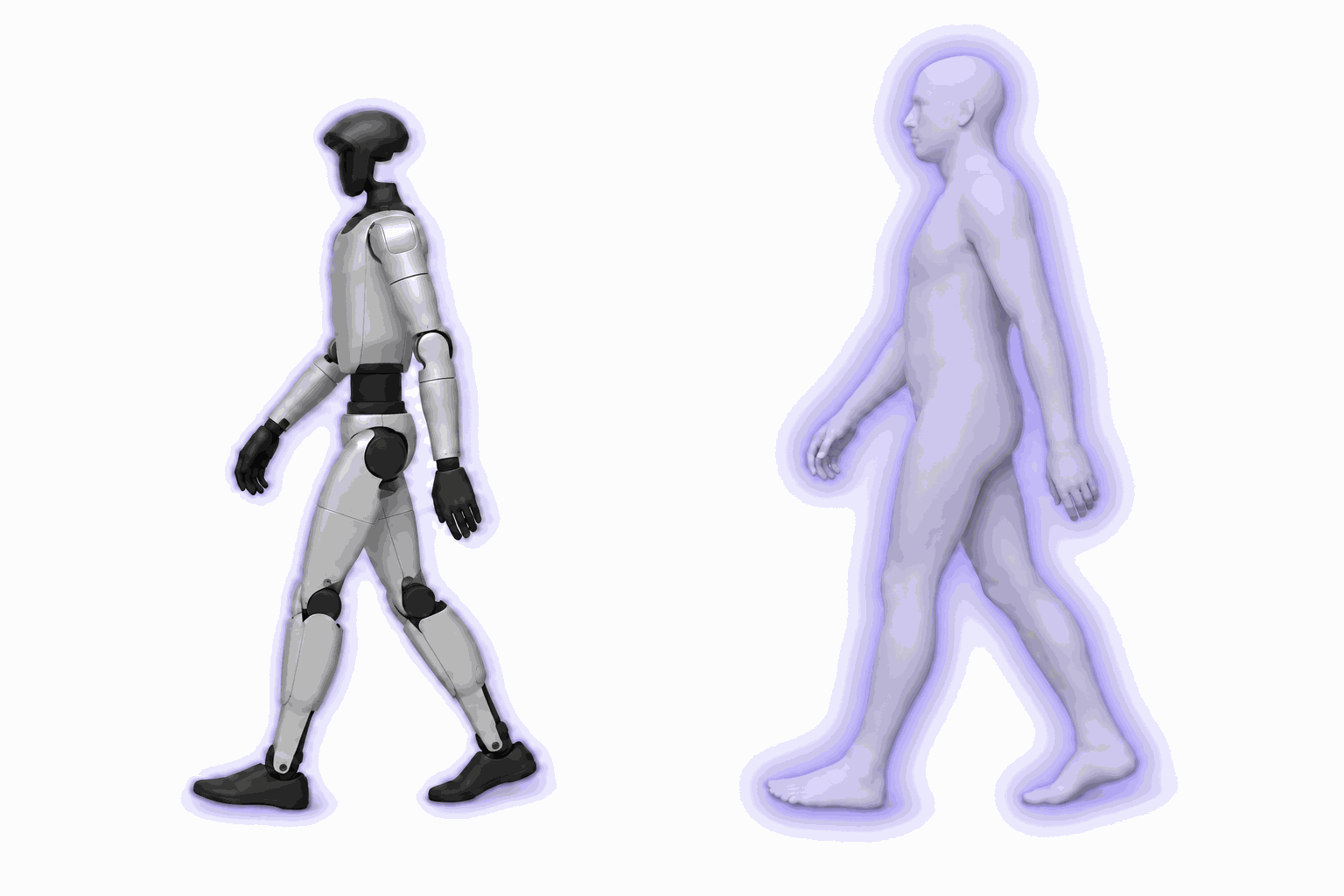}};
    \node[inner sep=0pt,opacity=.5] at (0,0) {\includegraphics[viewport=120 0 710 1010,clip,height=13pt]{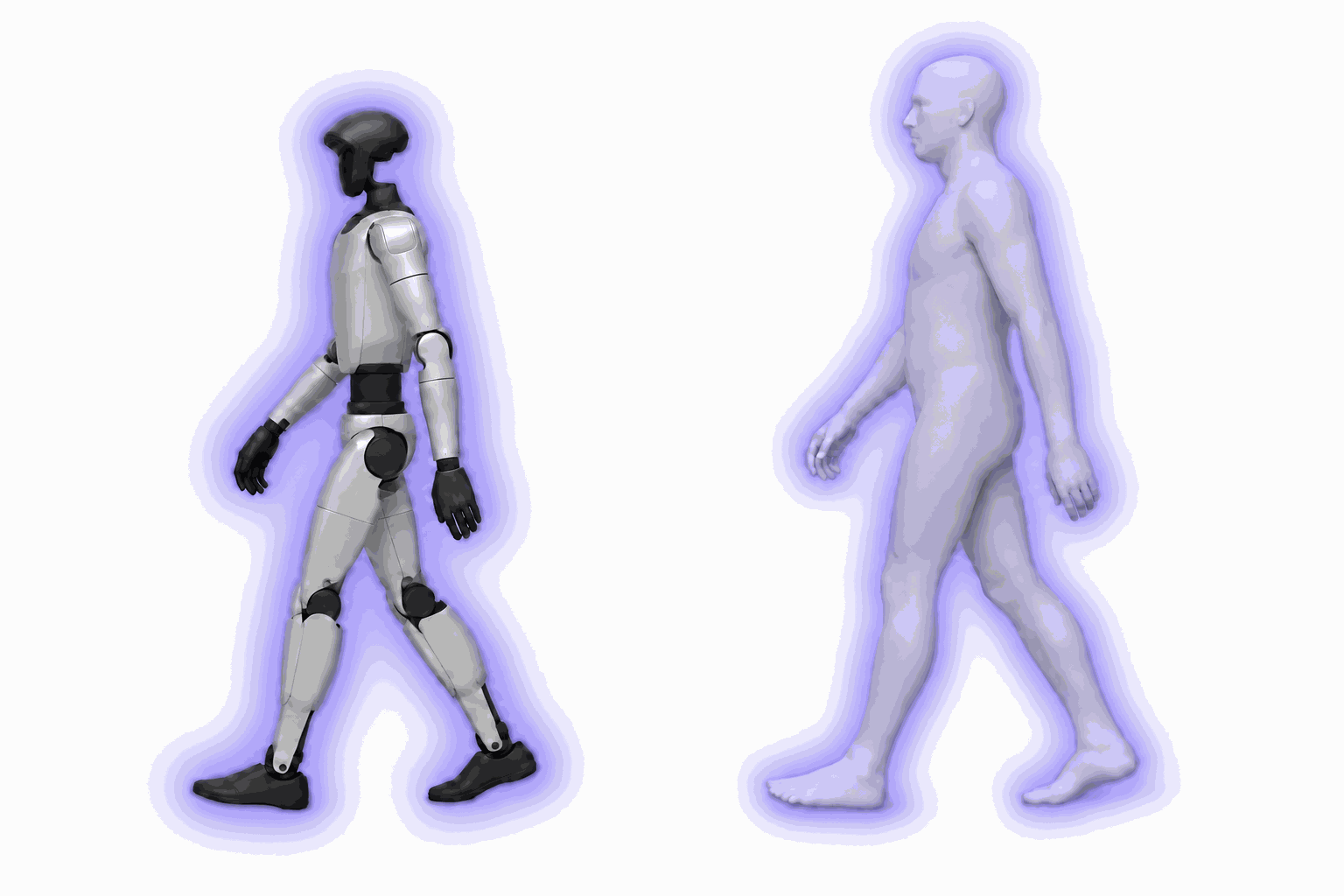}};
    \ifnum#1<3
      \draw[gray!65,line width=.3pt] (4.6,0) -- (6.6,0) (5.6,-1) -- (5.6,1);
      \node[inner sep=0pt] at (11,0) {\includegraphics[viewport=800 0 1380 1024,clip,height=12.2pt]{figs/generalization_strong_glow_icons.png}};
      \def\inlineobjectx{23}
      \draw[-{Stealth[length=1.5pt,width=1.3pt]},gray!65,line width=.3pt] (15.5,0) -- (19,0);
    \else
      \def\inlineobjectx{12}
      \draw[-{Stealth[length=1.5pt,width=1.3pt]},gray!65,line width=.3pt] (4.5,0) -- (8,0);
    \fi
    \begin{scope}[shift={(\inlineobjectx,0)},draw=referencepurple,line width=.4pt]
      \ifnum#1=1
        \path[draw,fill=referencepurple!12] (0,3) -- (2.5,1.5) -- (2.5,-1.5) -- (0,-3) -- (-2.5,-1.5) -- (-2.5,1.5) -- cycle;
        \draw (-2.5,1.5) -- (0,0) -- (2.5,1.5) (0,0) -- (0,-3);
        \fill[white] (2,-2) circle (1.5pt);
        \draw[line width=.5pt] (1,-2) -- (1.8,-2.8) -- (3.2,-1.1);
      \else
        \draw[dash pattern=on .8pt off .6pt] (0,3) -- (2.5,1.5) -- (2.5,-1.5) -- (0,-3) -- (-2.5,-1.5) -- (-2.5,1.5) -- cycle;
        \draw[referencepurple!50] (-2.5,1.5) -- (0,0) -- (2.5,1.5) (0,0) -- (0,-3);
      \fi
    \end{scope}
  \end{tikzpicture}%
}

\hypersetup{
  colorlinks=true,
  linkcolor=referencepurple,
  citecolor=referencepurple,
  urlcolor=referencepurple
}
\newif\ifshowrevisions
\showrevisionsfalse

\newcommand{\datasetname}{\textsc{HumanVerse}\xspace}
\newcommand{\modelname}{\ensuremath{\lambda_{0}}\xspace}

\title{$\lambda_0$: Scaling Open-World Egocentric Human Data for Humanoid Loco-Manipulation}

\DeclareRobustCommand{\affiliationlogo}[2][]{%
  \raisebox{-0.2\height}{\includegraphics[height=12pt,#1]{#2}}%
}

\DeclareRobustCommand{\teamlogo}{%
  \raisebox{-0.2\height}{\includegraphics[height=19pt,trim=70 100 60 75,clip]{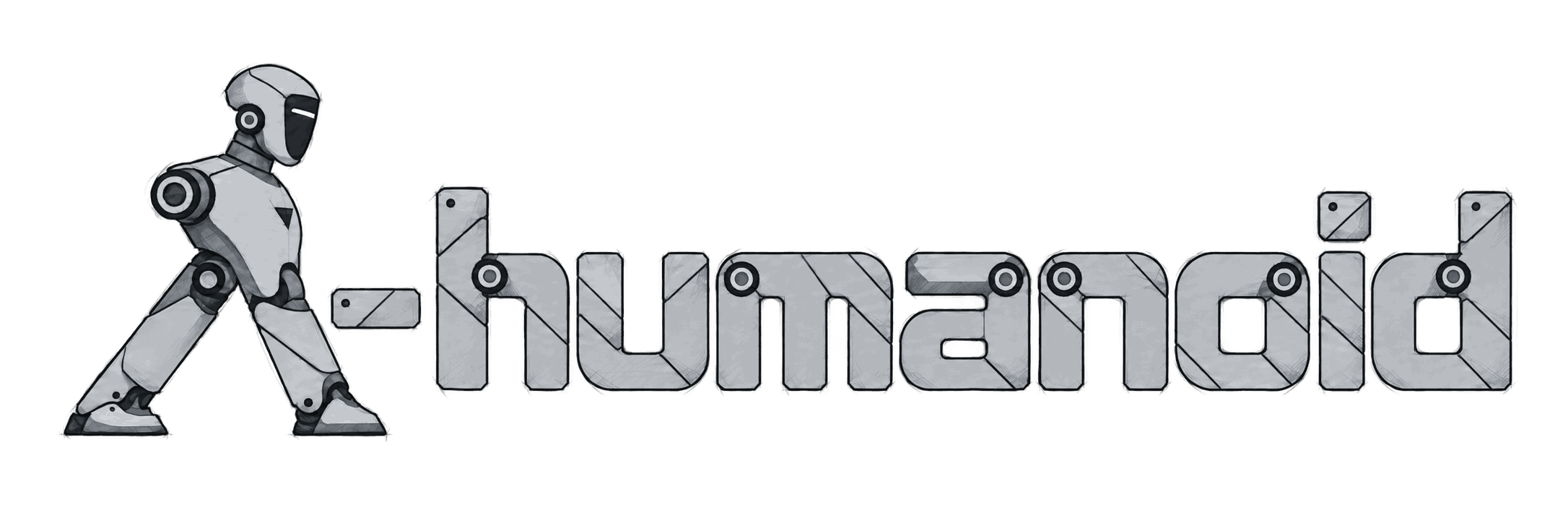}}%
}

\author{%
  \teamlogo\enspace Team\thanks{All contributors are listed in Sec.~\ref{sec:contributions}.}\\[6pt]
  Alibaba Token Hub\,\affiliationlogo[trim=0 0 430 0,clip]{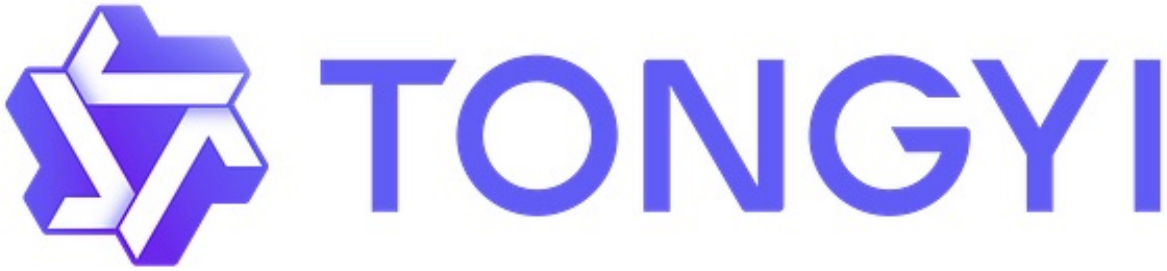} and  Amap\,\affiliationlogo{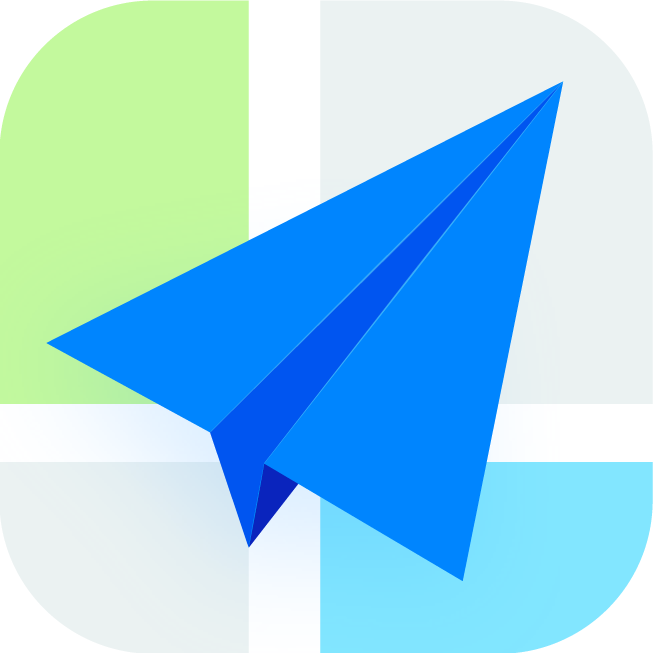}, Alibaba Group%
}

\begin{document}

\maketitle

\begin{figure}[!h]
  \centering
  \includegraphics[width=\linewidth]{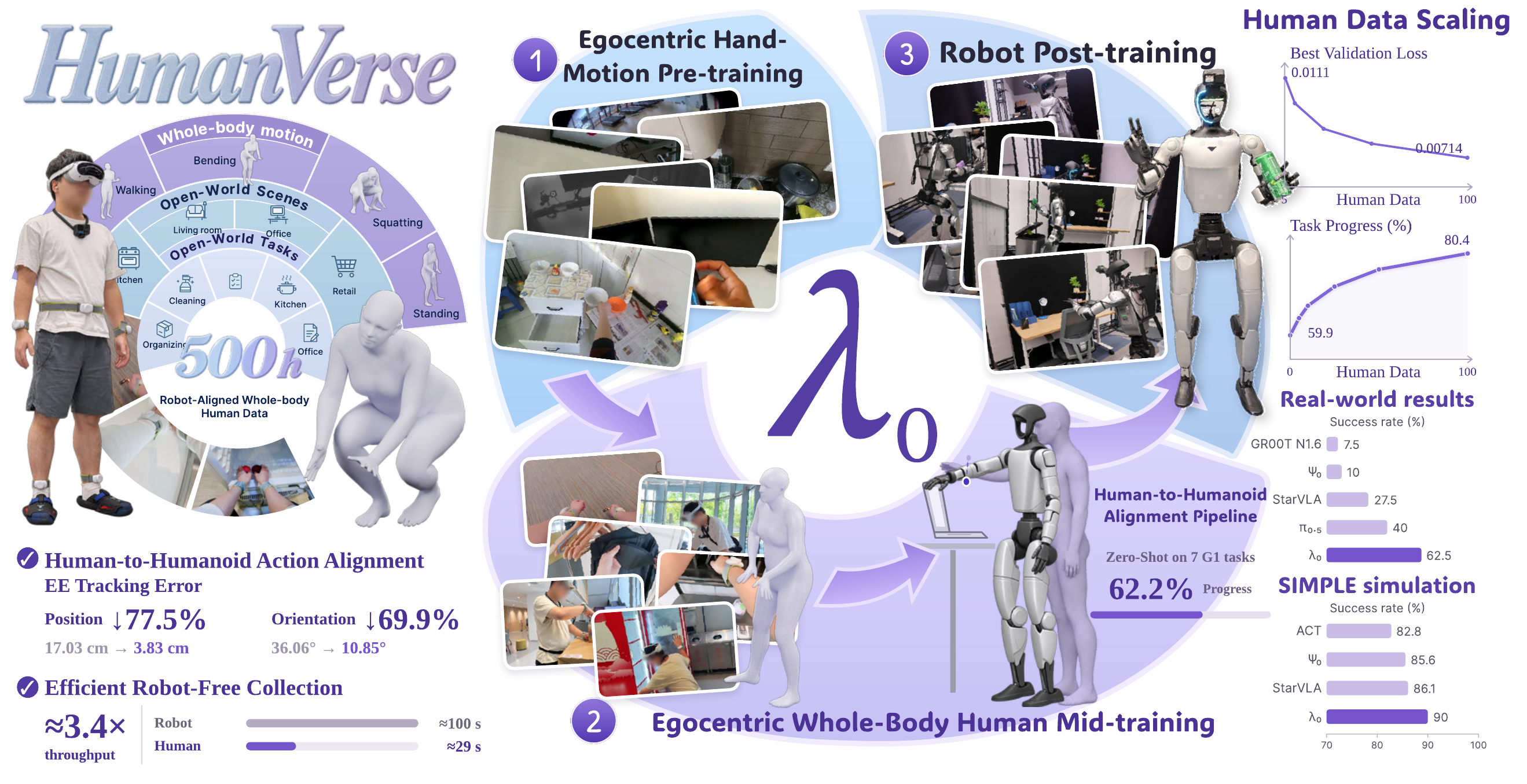}
  \caption{Introducing $\bm{\lambda_{0}}$ with \textbf{\textsc{HumanVerse}}.
  Efficient robot-free capture and human-to-humanoid alignment enable scalable whole-body human supervision, supporting three-stage training and improved humanoid loco-manipulation as human data scales.}
  \label{fig:teaser}
\end{figure}

\begin{abstract}
Humanoid loco-manipulation requires coordinated whole-body interaction with the physical world, yet collecting robot demonstrations at scale remains costly.
Human demonstrations offer a scalable alternative, but existing egocentric human supervision often omits locomotion and posture, while embodiment differences and controller dynamics hinder accurate end-effector execution.
We introduce \datasetname, a 500-hour dataset of open-world human loco-manipulation pairing egocentric video with synchronized body and hand motion.
Our human-to-humanoid pipeline aligns camera viewpoints and arm appearance, and refines motion references and controller inputs offline to reduce end-effector errors while preserving whole-body coordination.
We validate the pipeline through zero-shot transfer to Unitree G1 on representative whole-body manipulation and long-range loco-manipulation tasks, without target-task robot demonstrations.
Building on this pipeline, we develop \modelname\footnote{$\lambda$ resembles ``\begin{CJK*}{UTF8}{gkai}人\end{CJK*}'' (human in Chinese), symbolizing learning from everyday human experience.}, a humanoid foundation model for whole-body loco-manipulation, with robot-aligned human demonstrations providing supervision for mid-training.
Increasing whole-body human data consistently improves downstream task progress.
\modelname achieves 90.0\% success on SIMPLE and 62.5\% across real-world loco-manipulation tasks.
We expect this study to guide scalable whole-body human data collection toward humanoid foundation models for richer interaction with the physical world.

\end{abstract}

\begin{center}
\faGithub~Code: \href{https://github.com/LambdaHumanoid/openlambda}{\url{https://github.com/LambdaHumanoid/openlambda}}\\[2pt]
\textcolor{referencepurple}{\faHome}~Homepage: \href{https://lambdahumanoid.github.io}{\url{https://lambdahumanoid.github.io}}%
\end{center}

\clearpage
\startcontents[overview]
\begingroup
\hypersetup{linkcolor=black}
\edef\overviewtocdepth{\number\value{tocdepth}}
\setlength{\parskip}{0pt}
\small
\noindent{\large\bfseries Contents}\par\medskip
\titlecontents{overviewsection}[1.8em]
  {\addvspace{5pt}\small\bfseries}
  {\contentslabel{1.8em}}
  {\hspace*{-1.8em}}
  {\titlerule*[0.5pc]{.}\contentspage}
\titlecontents{overviewsubsection}[4.2em]
  {\small}
  {\contentslabel{2.4em}}
  {}
  {\titlerule*[0.5pc]{.}\contentspage}
\printcontents[overview]{overview}{1}{\setcounter{tocdepth}{2}}
\setcounter{tocdepth}{\overviewtocdepth}
\endgroup
\clearpage

\section{Introduction}
Learning humanoid loco-manipulation across tasks and environments requires demonstrations that capture coordinated locomotion, posture, and manipulation~\citep{jiang2026wholebodyvla,shi2026egohumanoid,wei2026psi0,hu2026openhlm,figure2026helix02,figure2026helix25}.
Whole-body teleoperation provides visual observations paired with robot actions, but requires operators to guide a physical humanoid through each demonstration~\citep{fu2025humanplus,he2025omnih2o,luo2025sonic,gear2025grootn16,li2026omega0}.
This dependence on robot availability and operation makes broader task and environment coverage costly to obtain~\citep{shi2026egohumanoid,punamiya2026egoverse}.
Scaling humanoid learning motivates a source of whole-body interaction experience that can grow beyond robot-operated collection.

Egocentric video records people's everyday interactions across tasks and environments without requiring robot operation~\citep{grauman2022ego4d,kareer2024egomimic,hoque2025egodex,punamiya2026egoverse,figure2026index}.
Recent methods recover wrist and hand trajectories from these videos, turning observed interactions into action supervision for robot learning~\citep{li2025vitra,yang2025egovla,wang2026ego2robot}.
Pretraining on these action-labeled videos can improve downstream manipulation~\citep{zheng2026egoscale,punamiya2026egoverse}.
For humanoid loco-manipulation, however, wrist and hand trajectories alone omit the locomotion and postural changes that support whole-body interaction~\citep{shi2026egohumanoid,zhao2026halomi}.
Recording synchronized body and hand motion captures this coordination, but motion retargeting and whole-body tracking can still leave end-effector errors due to geometric mismatch and imperfect execution~\citep{araujo2025gmr,luo2025sonic}.
These errors can compromise grasping and contact.
This raises our central question: \emph{Can egocentric whole-body human experience provide supervision that is both transferable to humanoids and practical to acquire at scale?}

We develop a human-to-humanoid alignment pipeline that captures human demonstrations without a robot and aligns their observations and actions to the downstream humanoid (Sec.~\ref{sec:human-robot-alignment}).
We match the camera and capture height to the robot setup, then remove human forearms and trackers from the images and composite rendered G1 arms.
Action alignment refines motion references and controller inputs offline to improve end-effector accuracy, producing robot-compatible state--action supervision.
We evaluate policies trained with this supervision on Unitree G1 across seven representative tasks spanning coordinated whole-body manipulation and long-range loco-manipulation (Sec.~\ref{sec:exp-zeroshot}).
Using task-specific human demonstrations without target-task robot demonstrations, the evaluations achieve mean progress of 51.7\% and 76.2\%, respectively.

Using this capture setup, we build \datasetname, a 500-hour corpus of human loco-manipulation spanning open-world scenes and tasks.
We record egocentric video and body motion with lightweight wearable sensors, synchronize the streams, and reconstruct hand motion from video.

We develop \modelname, a whole-body humanoid vision--language--action policy that learns from scalable human experience and progressively adapts it to humanoid control (Fig.~\ref{fig:teaser}).
Egocentric hand-motion pretraining draws on diverse human interaction videos.
Mid-training on \datasetname{} then provides a scalable, robot-free source of supervision for learning coordinated body and hand actions, bridging video pretraining and limited, costly robot demonstrations.
Robot post-training adapts the learned behaviors to the target humanoid.

Increasing whole-body human data reduces validation loss and raises real-world task progress monotonically from 59.9\% without whole-body mid-training to 80.4\% with the full corpus.
Ablations identify whole-body mid-training as the larger contributor to real-world success among the two human-data stages.
The full \modelname{} policy achieves 90.0\% success on SIMPLE and 62.5\% across four real-world loco-manipulation tasks, improving over the strongest evaluated baselines by 3.9\% and 22.5\%, respectively.
Human-data training also improves progress on robot-unseen object instances observed in human demonstrations.
Our contributions are:
\begingroup
\settowidth{\leftmargini}{\labelitemi}
\addtolength{\leftmargini}{\labelsep}
\begin{itemize}
  \item We develop a human-to-humanoid alignment pipeline for whole-body skill transfer and evaluate its collection efficiency and zero-shot transfer on G1 without target-task robot demonstrations.
  \item We introduce \datasetname, a 500-hour dataset pairing egocentric video with synchronized body and hand motion across open-world scenes and tasks.
  \item We develop \modelname{} to progressively adapt egocentric human interaction experience to humanoid control, and demonstrate that scaling human data with whole-body supervision improves downstream loco-manipulation.
\end{itemize}
\endgroup

\section{Related Work}
\subsection{Generalist Robot Policies}
Large-scale robot datasets combine demonstrations across tasks and embodiments, supporting the development of generalist policies~\citep{dasari2020robonet,jang2022bcz,walke2023bridgedata,openx2024,brohan2023rt1,zitkovich2023rt2,khazatsky2024droid,bu2025agibot,zhao2025humanoideveryday}.
These policies learn from diverse demonstrations within a shared model~\citep{ghosh2024octo,kim2025openvla,chi2025diffusion}, motivating more scalable ways to collect training data.
Advances in teleoperation and portable demonstration interfaces reduce collection effort~\citep{dass2023pato,qin2023anyteleop,zhao2023aloha,fu2025mobilealoha,chi2024umi}.
UMI on Legs extends robot-free hand-held demonstrations to quadruped manipulation by coupling a task-level manipulation policy with a simulation-trained whole-body controller~\citep{ha2024umionlegs}.
Robot-based collection depends on hardware availability and repeated resets, while portable interfaces have limited action spaces.
%

\subsection{Learning from Human Videos}
Human video datasets capture diverse activities, objects, and environments, providing rich experience for robot learning~\citep{grauman2022ego4d,damen2022epic,goyal2017something,grauman2024egoexo4d,wang2023holoassist,detone2026nymeriaplus,lightwheel2026egosuite}.
Learning from it requires connecting observations to actions.
Nymeria pairs synchronized multimodal egocentric recordings with full-body motion across 300 hours of daily activities in the wild~\citep{ma2024nymeria}, providing a foundation for studying context-dependent human motion.
Temporal and geometric annotations describe when interactions occur and how hands and objects move~\citep{liu2022hoi4d,wang2025hocap,sener2022assembly101,ragusa2021meccano,huang2024egoexolearn,banerjee2025hot3d}, while motion tracking and reconstruction provide human trajectories for action learning~\citep{hoque2025egodex,punamiya2026egoverse,li2025vitra,zheng2026egoscale,wang2026ego2robot}.
OpenEgo standardizes hand-pose annotations and temporally localized action descriptions across egocentric datasets~\citep{jawaid2025openego}.
Human videos also support visual representation learning~\citep{nair2023r3m,sontakke2023roboclip,kumar2020subroutines} and robot control through imitation and alignment~\citep{mandlekar2022offline,shao2020concept2robot,bahl2022whirl,wang2023mimicplay,bharadhwaj2023zeroshot,kareer2024egomimic,lepert2025phantom,yang2025egovla}.
HAT unifies human and humanoid state--action spaces for co-training on aligned PH2D demonstrations~\citep{qiu2025hat}, while DexWild combines wearable human interaction data with robot demonstrations for dexterous manipulation across environments~\citep{tao2025dexwild}.
EgoZero learns closed-loop manipulation from smart-glasses demonstrations without robot data using morphology-agnostic representations~\citep{liu2025egozero}, while HumanEgo uses hand--object interaction representations to learn robot-data-free policies from short human recordings~\citep{wang2026humanego}.
Beyond stationary manipulation, EMMA co-trains human full-body mobile manipulation demonstrations with static robot data, avoiding mobile robot teleoperation and reporting improved performance as human data increases~\citep{zhu2025emma}.
We use synchronized body and hand motion as action supervision for bipedal humanoid loco-manipulation, and measure the benefit of increasing this data while holding robot demonstrations fixed.

\subsection{Humanoid Whole-Body Control}
Humanoid whole-body control coordinates locomotion, posture, and manipulation to support complex physical interactions~\citep{fu2025humanplus,he2025omnih2o,luo2025sonic}.
Motion retargeting is a key interface: GMR studies how retargeting artifacts affect downstream tracking~\citep{araujo2025gmr}, and Human2Humanoid combines morphology-invariant trajectory alignment with physics-aware feasibility constraints~\citep{huang2026human2humanoid}.
Humanoid policies learn task-level behaviors from human experience through latent video actions, geometric pretraining, or joint training on human and robot demonstrations~\citep{jiang2026wholebodyvla,wei2026psi0,shi2026egohumanoid,hu2026openhlm,bjorck2025groot,beingbeyond2026beingh}.
ZeroWBC generates and tracks whole-body motion from an initial egocentric image and language instruction without robot teleoperation demonstrations~\citep{yang2026zerowbc}; HALOMI couples egocentric demonstration capture with view alignment and controller-aware reference adaptation for active-perception loco-manipulation~\citep{zhao2026halomi}.
Simulation provides another route from human video to control: VideoMimic reconstructs human motion and scenes for contextual control~\citep{allshire2025videomimic}, and SUGAR physically refines human-video interaction priors before distilling autonomous policies~\citep{wu2026sugar}.
GRAIL instead synthesizes interactions from 3D assets and video priors to train visual policies through simulation~\citep{xie2026grail}.
LEGS combines procedurally generated demonstrations with photorealistic Gaussian-splatting scene backgrounds to fine-tune humanoid loco-manipulation VLAs without teleoperation~\citep{kim2026legs}.
Predictive models and integrated whole-body systems offer complementary ways to connect perception, motion, and control~\citep{li2026omega0,symbiosis2026dpc,figure2026helix02,parada2026geminirobotics2,zheng2026motionwam,ye2026dreamzero,nvidia2026dreamdojo}.
Our alignment pipeline addresses both geometric mismatch in motion references and controller execution error, so that recorded human motions can supervise the action interface used by G1.

Supplementary Table~\ref{tab:closest-work} compares human-data interfaces, training strategies, and scaling studies.

\section{HumanVerse: Scaling Egocentric Whole-Body Data Collection in the Open World}
\label{sec:humanverse-data}
\begin{figure}[!t]
  \centering
  \includegraphics[width=\linewidth]{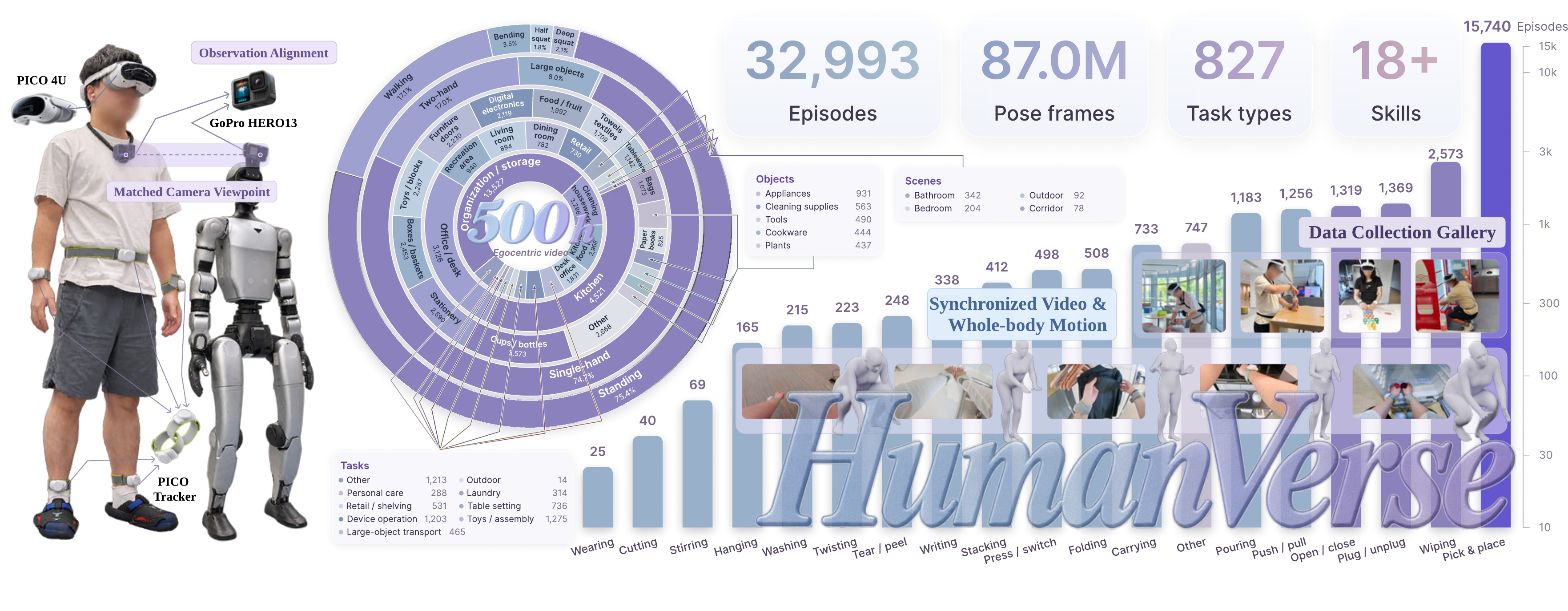}
  \caption{\textbf{\datasetname.}
  The wearable capture setup, task and motion distributions, and paired egocentric video and whole-body motions illustrate the dataset's scale and diversity.}
  \label{fig:hardware-setup}
\end{figure}

\datasetname contains 500 hours of egocentric human loco-manipulation, comprising 32,993 episodes and 87 million pose frames (Fig.~\ref{fig:hardware-setup}).
The dataset covers 827 task types across diverse objects and scenes, pairing first-person video with synchronized body and hand motion.
We convert these recordings into \emph{robot-aligned whole-body human data} for training; the 500-hour figure describes the recorded human corpus before conversion and validity filtering.

\subsection{Data Collection Setup and Protocol}
\label{sec:data-collection}

\begin{figure}[!ht]
  \centering
  \includegraphics[width=\linewidth]{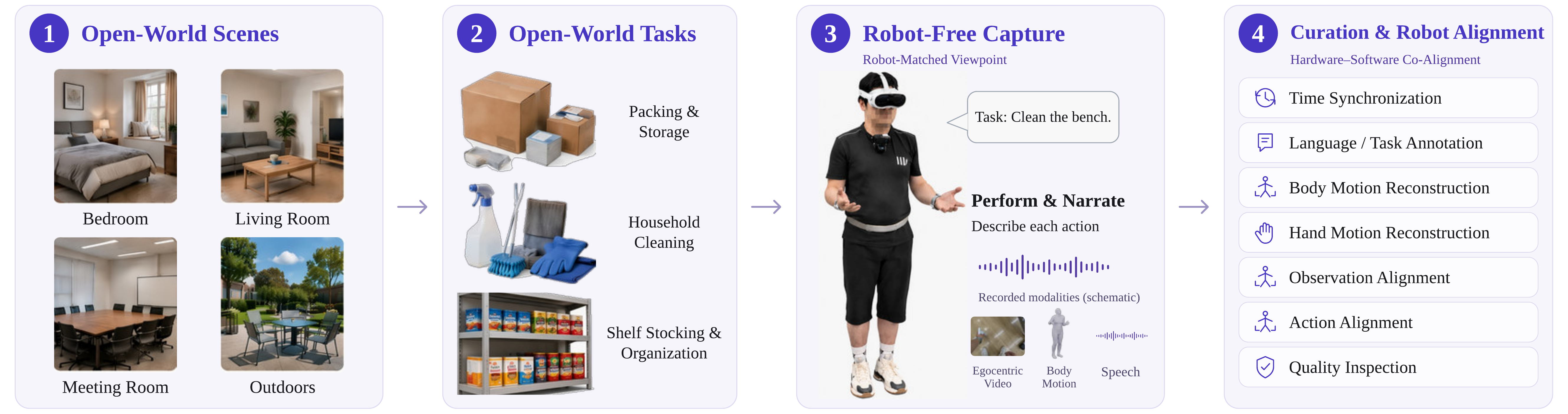}
  \caption{\textbf{Robot-free capture, curation, and robot alignment.}
  We select open-world scenes and tasks and record egocentric video, body motion, and speech with a robot-matched camera viewpoint. Curation includes synchronization, language/task annotation, body and hand motion reconstruction, observation alignment, action alignment, and quality inspection. Modality thumbnails are schematic.}
  \label{fig:data-collection-pipeline}
\end{figure}

We choose the GoPro HERO13 for convenient human data collection and equip our downstream robot, Unitree G1, with the same camera on its head.
We align the human camera height with the robot camera using a neck mount, while a PICO 4 Ultra headset and body trackers capture body motion alongside the video.
We reconstruct hand motion from the same video.
Figure~\ref{fig:data-collection-pipeline} summarizes the workflow from open-world scene and task selection to robot-free capture, data curation, and robot alignment.

\paragraph{Collection protocol.}
Collectors receive activity-level prompts rather than prescribed joint trajectories, allowing variation in walking paths, initial posture, object instances, placements, grasp strategies, and intermediate corrections.
Each episode preserves the approach, repositioning, and transitions needed to complete an interaction, rather than only the final hand motion.
Collectors use the wearable interface in everyday locations without an on-site robot, focusing on carrying, tidying, cleaning, and operating objects and fixtures.
We retain original streams, collector and session identifiers, activity labels, calibration, and device-health records. On-site checks flag interruptions, missing modalities, and calibration failures.

\subsection{Dataset Scope and Coverage}
\label{sec:dataset-scope}

\datasetname denotes only the 500-hour synchronized egocentric whole-body corpus. External hand--object datasets and robot demonstrations are excluded from its duration and scaling subsets; their training roles are described in Sec.~\ref{sec:training-recipe}.

\datasetname captures everyday activities in homes, retail and supermarket spaces, industrial sites, offices, and outdoor areas. Activities include cleaning, cooking, personal care, laundry, table setting, and large-object transport.
For example, interacting with objects at different heights involves bending or reaching, while transporting objects couples walking with sustained grasping.
Figure~\ref{fig:hardware-setup} summarizes the task, scene, object, and motion distributions, while Supplementary Table~\ref{tab:egocentric-datasets} compares the dataset's capture setup and motion coverage with existing egocentric datasets.

\begin{figure}[!htbp]
  \centering
  \includegraphics[width=\linewidth]{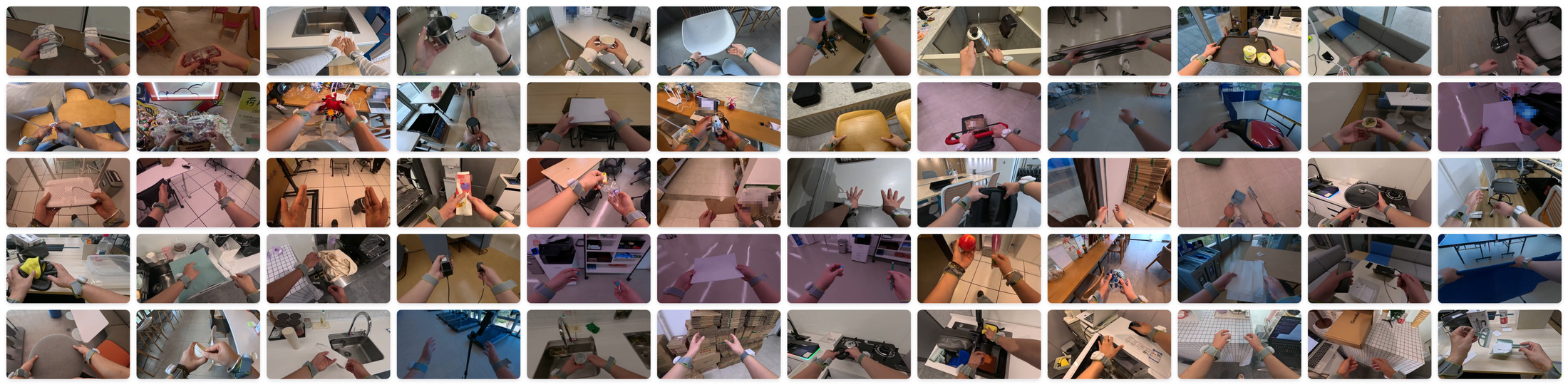}
  \caption{\textbf{Representative examples from \datasetname.}
  Sixty first-person frames are arranged in a $12\times5$ gallery to illustrate diversity in tasks, scenes, objects, and hand--object interactions.}
  \label{fig:data-diversity-examples}
\end{figure}

\paragraph{Qualitative task diversity.}
Figure~\ref{fig:data-diversity-examples} presents 60 egocentric frames illustrating everyday tasks, scenes, and manipulated objects.
Each tile depicts one moment from a capture session.
These examples illustrate qualitative diversity and do not estimate dataset-wide task frequencies.

\subsection{Human Motion Reconstruction and Annotation}
\label{sec:motion-reconstruction}

\paragraph{Temporal synchronization and language annotation.}
The GoPro and PICO run on independent clocks. We assign each episode a time offset under the convention $t_{\mathrm{pico}}=t_{\mathrm{gopro}}+\Delta$ and refine a capture-time prior by cross-correlating GoPro gyroscope measurements with tracker-derived body angular velocity.
After applying this offset, video and body pose are resampled onto a common timeline over their overlapping interval, with synchronization quality retained as an explicit diagnostic.
Spoken instructions on the GoPro audio track are transcribed into episode-level language/task labels.
The records retain stable frame indices for reconstruction and alignment.

\paragraph{Body and hand motion.}
We reconstruct a synchronized human motion sequence from wearable body tracking and egocentric video. The PICO system~\citep{pico_body_tracking} supplies timestamped 24-joint body kinematics, while Dyn-HaMR~\citep{yu2025dynhamr} estimates 21 joints per hand using camera motion from ViPE~\citep{huang2025vipe}. Coordinate transforms register these trajectories to the synchronized video frames. These body and hand trajectories provide the reference for robot alignment in Sec.~\ref{sec:human-robot-alignment}.

\begin{figure}[!t]
  \centering
  \includegraphics[width=\linewidth]{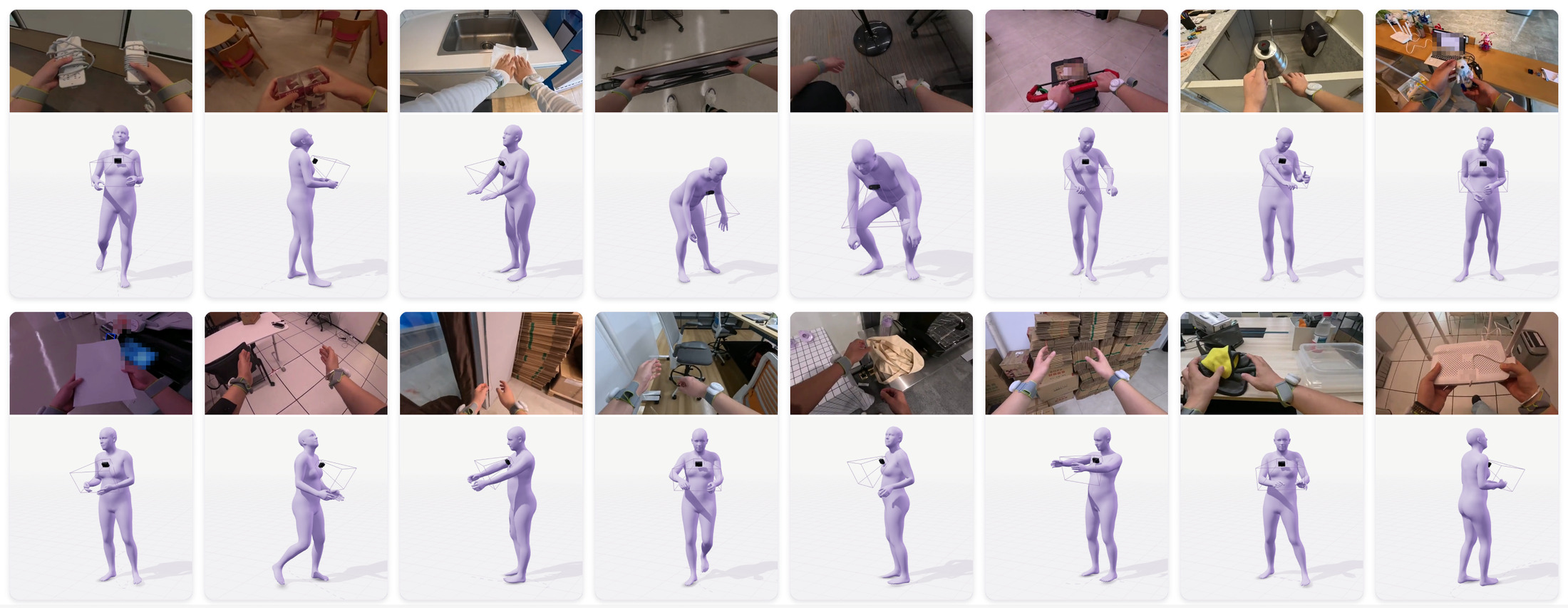}
  \caption{\textbf{Synchronized human motion and egocentric observations.}
  Top: egocentric RGB frames. Bottom: corresponding SMPL body meshes with MANO hand motion and illustrative camera frusta.
  Camera frusta schematically illustrate viewing direction.}
  \label{fig:smpl-camera-examples}
\end{figure}

\paragraph{Coordinate representations.}
We retain the camera trajectory and available camera-to-world and body-frame transforms. Hand trajectories are stored both in current-camera coordinates for reprojection and in an episode-fixed frame anchored at the first valid camera pose to preserve displacement over time. Finger geometry is represented in wrist-local, palm-aligned, scale-normalized coordinates.

\paragraph{Annotation records.}
Each derived motion sequence retains source-frame indices, coordinate transforms, conversion versions, per-track visibility, and estimator confidence. Raw recordings are the canonical source. Separate body and hand validity masks preserve usable body motion when hands are missing or occluded, without interpolating missing poses. The body stream consists of wearable-device kinematics; the hand trajectories are reconstructed from RGB. Validation and splits are detailed in Appendix~\ref{sec:dataset-validation}.

\paragraph{Data quality and splits.}
We partition \datasetname by recording session before extracting temporal windows and draw nested scaling subsets only from the frozen general-purpose training partition. Task-specific human demonstrations for downstream post-training remain separate from this corpus, and human-data selection does not consult the humanoid evaluation manifest. Appendix~\ref{sec:dataset-validation} details the quality checks, validity masking, and split protocol.

Figure~\ref{fig:smpl-camera-examples} shows 16 examples pairing egocentric frames with their reconstructed motions.

\subsection{Human-to-Humanoid Alignment Pipeline}
\label{sec:human-robot-alignment}

We align the recorded images and reconstructed motions to G1 (Fig.~\ref{fig:data-collection-pipeline}). Observation alignment matches camera settings and replaces human arms with rendered robot arms. Action alignment refines motion references and controller inputs offline, producing humanoid states and motion tokens in the action interface used at deployment.

\begin{figure}[!b]
  \centering
  \includegraphics[width=\linewidth]{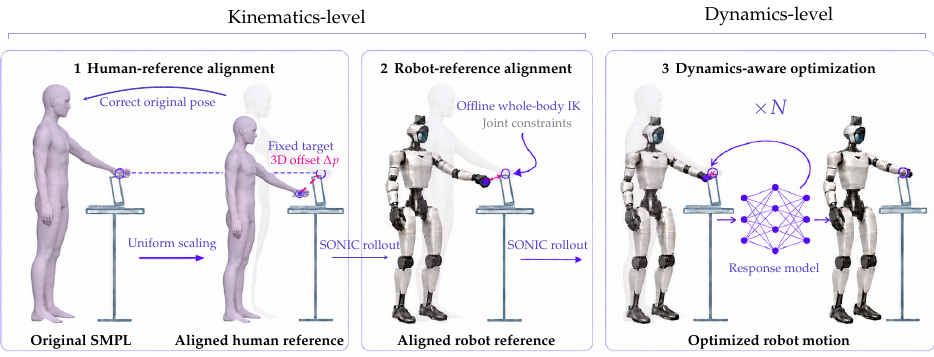}
  \caption{\textbf{Human-to-humanoid action alignment.}
  The first two stages modify the human and robot references. The third accounts for execution dynamics and directly optimizes motion tokens using gradient-based updates. End-effector targets remain fixed throughout. Rendered poses illustrate the procedure and are not experimental rollouts.}
  \label{fig:three-stage-alignment}
\end{figure}

\begingroup
\clubpenalties 4 10000 10000 10000 0
\paragraph{Action Alignment.}

Humanoids rely on a tracking policy such as SONIC~\citep{luo2025sonic} for coordinated whole-body loco-manipulation. Motion references obtained through retargeting, such as GMR~\citep{araujo2025gmr}, or human-motion execution through SONIC can still yield end-effector pose errors. Geometric mismatch affects the reference, while controller tracking introduces additional execution error. These errors can compromise grasping and contact. We refine the human reference, robot reference, and controller inputs offline in three stages (Fig.~\ref{fig:three-stage-alignment})~\citep{jiang2026egoalign,chen2026egohumanoidv2}. (1)~\textbf{Human-reference alignment:} Height differences between the human reference and G1 can shift end effectors away from the targets. We scale a SMPL copy to G1 height, adjust its pose to align end effectors with the fixed targets, and apply these corrections to the human reference. A SONIC rollout yields better-aligned G1 motion. (2)~\textbf{Robot-reference alignment:} The scaled SMPL skeleton still differs from G1's body geometry, leaving residual end-effector errors. We correct the robot reference through constrained offline whole-body inverse kinematics (IK) using G1's kinematic model. Re-encoding the corrected reference and replaying its tokens further reduces these errors. (3)~\textbf{Dynamics-aware optimization:} Execution dynamics introduce errors that kinematic correction alone cannot resolve. We use rollout feedback and a response model to guide gradient-based updates to the motion tokens. Multiple rounds of token updates and rollouts progressively reduce end-effector errors. During dataset production, we select the best valid result across the three stages. A SONIC rollout of the original human poses serves as a fallback if alignment yields no improvement. We validate the latter two stages on 50 tasks (about 1~h of data), reducing task-averaged end-effector errors from 17.03 to 3.83~cm in position and from 36.06$^\circ$ to 10.85$^\circ$ in orientation.

\endgroup

\paragraph{Observation Alignment.}
We use the same camera model and match capture height to reduce viewpoint differences between human and robot recordings. To align arm appearance, we mask human forearms and wrist-mounted PICO trackers with SAM2~\citep{ravi2024sam2} and inpaint these regions with DiffuEraser~\citep{li2025diffueraser}. We then render G1 arms from the aligned joint positions, fit them to the human wrists and forearms, and composite them onto the inpainted frames. This preserves the scene and interaction. Human and humanoid motion can still alter camera viewpoints. During training, random resized crops, small rotations, and photometric jitter improve robustness to these differences. The deployed policy uses humanoid images.

\section{\texorpdfstring{$\lambda_0$}{lambda0}: Learning Humanoid Loco-manipulation from Human Experience}
\label{sec:method}
\begin{figure*}[t]
  \centering
  \includegraphics[width=\textwidth]{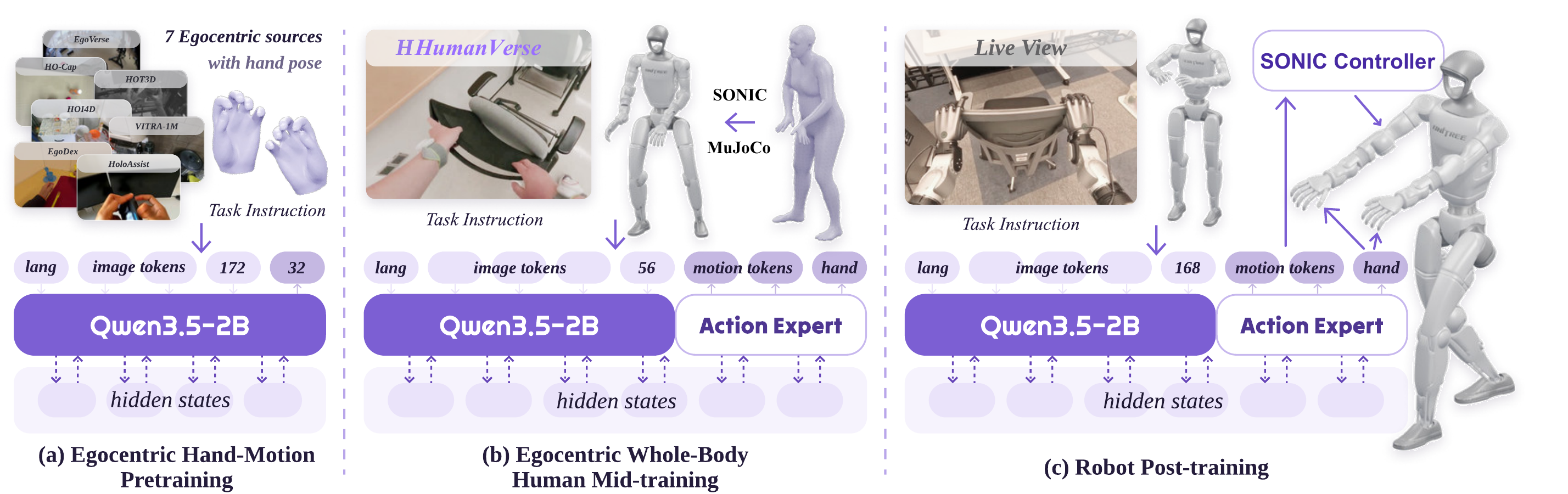}
  \caption{\textbf{Three-stage training of \modelname.}
  The policy learns interaction from egocentric data, whole-body coordination from \datasetname, and downstream control from real-robot data.}
  \label{fig:training-recipe}
\end{figure*}

\modelname uses a vision--language--action architecture and domain-specific interfaces to learn task-conditioned actions from human experience. Offline action alignment improves the end-effector execution accuracy of human-derived motions; VLA training learns to predict actions from observations, language, and state using the resulting supervision.
Training proceeds through egocentric hand-motion pretraining, egocentric whole-body human mid-training, and robot post-training (Fig.~\ref{fig:training-recipe}).
Section~\ref{sec:human-robot-alignment} supplies Stage II with aligned human-derived states and actions in the same motion-token interface used by the deployed whole-body controller. Stage III then adapts the policy with robot data.

\subsection{Policy Architecture and Action Interfaces}
\label{sec:state-action-interfaces}

\paragraph{Architecture and inputs.}
\modelname combines a Qwen3.5-2B~\citep{qwen2026qwen35} vision--language backbone with a generative action expert. In Stages II and III, the policy takes an egocentric image, a language instruction, and the current state. For G1, the 44-dimensional state contains 29 body-joint positions, six hand commands per side, and three projected-gravity components. Following $\pi_{0.5}$~\citep{physicalintelligence2025pi05}, we discretize the state and append it to the instruction. Layer-wise backbone features condition the expert.

\paragraph{Action interfaces.}
The action expert predicts a 50-step action chunk; for G1, each step contains a 64-dimensional SONIC motion latent and two 6-dimensional Revo2 hand commands, totaling 76 dimensions. Stage I instead predicts future discrete bilateral hand-state tokens using only the backbone. The action expert is introduced in Stage II and shared with Stage III. Human-derived mid-training data and native G1 post-training data use the same state--action format (Appendix~\ref{sec:app-action-interfaces}) but distinct domain identifiers and input/output projections. Following X-VLA~\citep{zheng2025xvla}, domain-specific projections and learned soft prompts support cross-domain transfer.

\subsection{Three-Stage Training}
\label{sec:training-recipe}
\label{sec:training-data}

Our training pipeline uses three data sources: public egocentric hand--object datasets ($\mathcal{D}_E$) for pretraining on egocentric videos and hand poses, \datasetname{} ($\mathcal{D}_W$) for mid-training on whole-body human motion, and robot demonstrations ($\mathcal{D}_R$) for robot post-training.

Stage I predicts future hand-state tokens in the vision--language backbone. Stages II and III use the masked velocity-field loss $\mathcal{L}^{q}$ defined in Sec.~\ref{sec:training-objective-deployment}, with different data sources and active action interfaces.
Let $\boldsymbol{\phi}^{(1)}$ denote the Stage-I backbone parameters and $\boldsymbol{\theta}^{(k)}$ the action-policy parameters learned in Stages II and III.
Stage II initializes its backbone from the Stage-I EMA checkpoint and introduces the action expert. Stage III initializes from the Stage-II EMA checkpoint.

\paragraph{Stage I: egocentric hand-motion pretraining.}
We initialize the backbone from a vision--language checkpoint and train it to predict a future sequence of discrete bilateral hand-state tokens from egocentric observations, the current hand state, and language (Fig.~\ref{fig:training-recipe}(a)). No action expert is trained.
Training uses EgoDex, EgoVerse, HO-Cap, HOI4D, HoloAssist, HOT3D, and VITRA-1M \citep{hoque2025egodex,punamiya2026egoverse,wang2025hocap,liu2022hoi4d, wang2023holoassist,banerjee2025hot3d,li2025vitra}, with hand-pose supervision from either native ground-truth annotations or pseudo-labels generated by open-source methods.
Let $\mathcal{S}_E$ denote these seven data sources. We sample datasets in proportion to the square root of their usable frame counts $N_d$ to increase the relative contribution of smaller sources:
\begin{equation}
  p_1(d)=\frac{\sqrt{N_d}}{\sum_{d'\in\mathcal{S}_E}\sqrt{N_{d'}}},
  \qquad
  \boldsymbol{\phi}^{(1)}=
  \arg\min_{\boldsymbol{\phi}}
  \mathbb{E}_{d\sim p_1,\,\tau\sim\mathcal{D}_d}
  \left[\mathcal{L}_{\mathrm{token}}^{E}(\tau;\boldsymbol{\phi})\right].
  \label{eq:stage1-main}
\end{equation}
Here $\mathcal{L}_{\mathrm{token}}^{E}$ is the future hand-state token prediction loss. Each bilateral hand state has 138 continuous dimensions: 3D wrist position, 6D rotation, and $20\times3$ local keypoint coordinates per hand.

\paragraph{Stage II: egocentric whole-body human mid-training.}
Stage II uses \datasetname{} demonstrations converted into the G1 state--action format by the pipeline in Sec.~\ref{sec:human-robot-alignment} (Fig.~\ref{fig:training-recipe}(b)). The supervision remains human-derived, with the domain-specific interfaces described above.
Denoting the conversion by $\Phi_R$ and the human-domain loss by $\mathcal{L}^{H}$, we initialize the backbone from Stage I, introduce the action expert and human-data projections, and optimize
\begin{equation}
  \boldsymbol{\theta}^{(2)}=
  \arg\min_{\boldsymbol{\theta}:\,\boldsymbol{\phi}\leftarrow
  \operatorname{EMA}(\boldsymbol{\phi}^{(1)})}
  \mathbb{E}_{\tau\sim\Phi_R(\mathcal{D}_W)}
  [\mathcal{L}^{H}(\tau;\boldsymbol{\theta})].
  \label{eq:stage2-main}
\end{equation}
We update the backbone, action expert, and human-data projections during this stage.

\paragraph{Stage III: robot post-training.}
We post-train the policy on robot demonstrations to adapt the motion and interaction patterns learned in the preceding stages to downstream tasks and executable robot control (Fig.~\ref{fig:training-recipe}(c)).
The post-training objective is
\begin{equation}
  \boldsymbol{\theta}^{(3)}=
  \arg\min_{\boldsymbol{\theta}\leftarrow
  \operatorname{EMA}(\boldsymbol{\theta}^{(2)})}
  \mathbb{E}_{\tau\sim\mathcal{D}_R}
  [\mathcal{L}^{R}(\tau;\boldsymbol{\theta})].
  \label{eq:stage3-main}
\end{equation}
We jointly optimize the model, embodiment conditioning, and active interfaces. Appendix~\ref{sec:app-stage-recipe} summarizes the training schedule.

\subsection{Training Objective and Deployment}
\label{sec:training-objective-deployment}

\paragraph{Training objective for Stages II and III.}
Stages II and III use flow matching following $\pi_{0.5}$~\citep{physicalintelligence2025pi05}.
For domain $q\in\{H,R\}$, let $\mathbf{a}$ be a target action chunk, $\mathbf{c}_{\tau}$ the observation, instruction, and state conditioning, $\boldsymbol{\epsilon}\sim\mathcal{N}(\mathbf{0},\mathbf{I})$, and $s\sim\mathcal{U}(0,1)$. On predicted entries, the noisy action is $\mathbf{x}_s=(1-s)\boldsymbol{\epsilon}+s\mathbf{a}$ and the target velocity is $\mathbf{a}-\boldsymbol{\epsilon}$. With $m^q_{h,j}=1$ only for valid, unprefixed scalar targets at step $h$ and dimension $j$, the masked velocity-field loss is
\begin{equation}
  \mathcal{L}^{q}(\tau;\boldsymbol{\theta})=
  \mathbb{E}_{s,\boldsymbol{\epsilon}}
  \left[
    \frac{\sum_{h,j}m^q_{h,j}\bigl(v^q_{\boldsymbol{\theta},h,j}(\mathbf{x}_s,s;\mathbf{c}_{\tau})-(a_{h,j}-\epsilon_{h,j})\bigr)^2}
    {\sum_{h,j}m^q_{h,j}}
  \right],
  \qquad q\in\{H,R\}.
  \label{eq:masked-velocity-loss}
\end{equation}
Provided action-prefix entries remain fixed during noising and are excluded from the loss; the mask also removes missing annotations and padding.
Inference uses 10 Euler steps from Gaussian noise to actions, keeping any provided action prefix fixed. See Appendix~\ref{sec:app-layerwise-policy} for details.

\paragraph{Deployment.}
At deployment, \modelname{} uses G1's egocentric image, language instruction, and proprioceptive state with the robot-domain identifier and projections (Fig.~\ref{fig:training-recipe}(c)).
The predicted 64-dimensional motion token drives the SONIC whole-body controller, and the two 6-dimensional hand commands drive the Revo2 hands. Deployment requires no human motion streams.

\section{Experiments on Humanoid: Alignment, Scale, and Transfer}
\label{sec:experiments}
\label{sec:evaluation-protocol}

We evaluate aligned human demonstrations without target-task robot data, then compare \modelname{} with policy baselines, vary human-data scale and training stages, and test transfer to object instances absent from robot demonstrations.

\subsection{Effectiveness and Efficiency of Robot-Aligned Human Data}
\label{sec:exp-zeroshot}

We assess robot-aligned human supervision along two dimensions: the behaviors it enables without target-task robot demonstrations, and the collection effort needed to learn them. Our evaluations cover whole-body manipulation and long-range loco-manipulation~\citep{chen2026egohumanoidv2,jiang2026egoalign}.

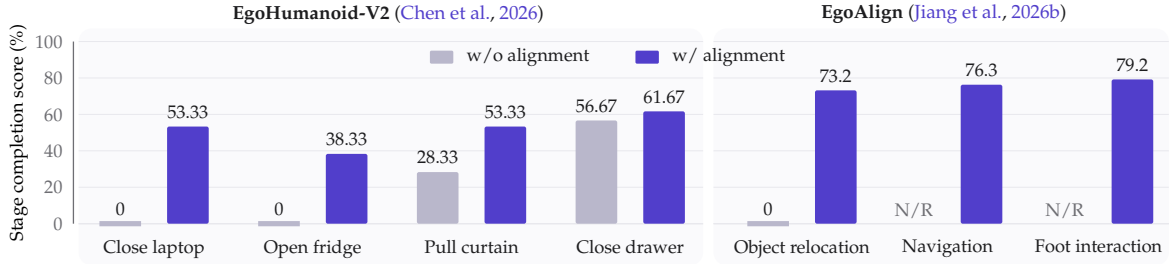
\begin{figure}[!htbp]
  \centering
  \resizebox{\linewidth}{!}{
\begingroup
\definecolor{alignink}{HTML}{1D1D1F}
\definecolor{alignsecondary}{HTML}{6E6E73}
\definecolor{alignrule}{HTML}{E8E8ED}
\definecolor{alignbefore}{HTML}{B9B7CB}
\definecolor{aligncard}{HTML}{FAFAFC}
\begin{tikzpicture}[x=1pt,y=1pt,font=\rmfamily\fontsize{7}{8}\selectfont]
  \path[use as bounding box] (0,-20) rectangle (450,90);
  \fill[aligncard,rounded corners=5pt] (30,-16) rectangle (270,75);
  \fill[aligncard,rounded corners=5pt] (274,-16) rectangle (449,75);
  \node[text=alignink,font=\rmfamily\fontsize{7.3}{8}\selectfont] at (149,80.5)
    {\textbf{EgoHumanoid-V2}~\citep{chen2026egohumanoidv2}};
  \node[text=alignink,font=\rmfamily\fontsize{7.3}{8}\selectfont] at (362.5,80.5)
    {\textbf{EgoAlign}~\citep{jiang2026egoalign}};
  \fill[alignbefore,rounded corners=.6pt] (165,61.5) rectangle (174,66.5);
  \node[anchor=west,inner sep=0pt,text=alignink] at (179,64) {w/o alignment};
  \fill[referencepurple,rounded corners=.6pt] (244,61.5) rectangle (253,66.5);
  \node[anchor=west,inner sep=0pt,text=alignink] at (258,64) {w/ alignment};
  \node[rotate=90,font=\rmfamily\fontsize{7}{8}\selectfont] at (7,35)
    {Stage completion score (\%)};
  \foreach \tick in {0,20,40,60,80,100}{
    \draw[alignrule,line width=.35pt] (30,0.7*\tick) -- (270,0.7*\tick);
    \draw[alignrule,line width=.35pt] (274,0.7*\tick) -- (449,0.7*\tick);
    \node[anchor=east,inner sep=1pt,text=alignsecondary] at (25,0.7*\tick) {\tick};
  }
  \draw[alignsecondary,line width=.5pt] (28,0) -- (28,70);
  \draw[alignrule,line width=.5pt] (30,0) -- (270,0);
  \draw[alignrule,line width=.5pt] (274,0) -- (449,0);
  \foreach \cx/\before/\after/\task in {
    59/0/53.33/{Close laptop},
    120/0/38.33/{Open fridge},
    181/28.33/53.33/{Pull curtain},
    242/56.67/61.67/{Close drawer},
    308/0/73.20/{Object relocation},
    364/-1/76.30/{Navigation},
    422/-1/79.20/{Foot interaction}}{
    \ifdim\before pt<0pt
      \node[text=alignsecondary,inner sep=0pt] at ({\cx-13},5) {N/R};
    \else
      \ifdim\before pt=0pt
        \draw[alignbefore,line width=2pt] ({\cx-21},0) -- ({\cx-5},0);
      \else
        \fill[alignbefore,rounded corners=.8pt]
          ({\cx-21},0) rectangle ({\cx-5},{.7*\before});
      \fi
      \node[anchor=south,inner sep=1pt,text=alignink] at ({\cx-13},{.7*\before+2})
        {\pgfmathprintnumber[fixed,precision=2,zerofill=false]{\before}};
    \fi
    \fill[referencepurple,rounded corners=.8pt]
      ({\cx+5},0) rectangle ({\cx+21},{.7*\after});
    \node[anchor=south,inner sep=1pt,text=alignink] at ({\cx+13},{.7*\after+2})
      {\pgfmathprintnumber[fixed,precision=2,zerofill=false]{\after}};
    \node[anchor=north,align=center,inner sep=0pt,text=alignink,
      font=\rmfamily\fontsize{7}{8}\selectfont] at (\cx,-6) {\task};
  }
\end{tikzpicture}
\endgroup}
  \caption{\textbf{Task progress from aligned human demonstrations.}
  Scores include partial completion; long-range results average seen and unseen target locations. N/R indicates missing before-alignment scores.}
  \label{fig:zeroshot-alignment}
\end{figure}

\begin{figure}[!ht]
  \centering
  \resizebox{0.84\linewidth}{!}{\begingroup
\definecolor{costrobot}{HTML}{527DAF}
\definecolor{costmuted}{HTML}{656575}
\definecolor{costgrid}{HTML}{E8E8ED}
\begin{tikzpicture}[x=1pt,y=1pt,font=\rmfamily\fontsize{8}{9}\selectfont]
\path[use as bounding box] (-7,-8.8) rectangle (354.75,106.48);
\draw[referencepurple,dashed,line width=1.2pt] (52.02,96.8)--(63.036,96.8);
\filldraw[fill=white,draw=referencepurple] (57.528,96.8) circle (2pt);
\node[anchor=west] at (65.484,96.8) {Human-only};
\draw[costrobot,line width=1.2pt] (135.864,96.8)--(146.88,96.8);
\fill[costrobot] (140.148,95.04) rectangle (142.596,98.56);
\node[anchor=west] at (149.328,96.8) {Robot-only};
\node[rotate=90] at (-1.104,49.28) {Mean task score (\%)};

\draw[costgrid,line width=.35pt] (22.644,17.6)--(214.2,17.6);\node[anchor=east,text=costmuted] at (20.808,17.6) {0};
\draw[costgrid,line width=.35pt] (22.644,33.44)--(214.2,33.44);\node[anchor=east,text=costmuted] at (20.808,33.44) {20};
\draw[costgrid,line width=.35pt] (22.644,49.28)--(214.2,49.28);\node[anchor=east,text=costmuted] at (20.808,49.28) {40};
\draw[costgrid,line width=.35pt] (22.644,65.12)--(214.2,65.12);\node[anchor=east,text=costmuted] at (20.808,65.12) {60};
\draw[costgrid,line width=.35pt] (22.644,80.96)--(214.2,80.96);\node[anchor=east,text=costmuted] at (20.808,80.96) {80};
\draw[costmuted,line width=.4pt] (22.644,80.96)--(22.644,17.6)--(214.2,17.6);
\node[anchor=north,text=costmuted,inner sep=2pt] at (29.376,15.84) {6};
\node[anchor=north,text=costmuted,inner sep=2pt] at (64.0958,15.84) {12};
\node[anchor=north,text=costmuted,inner sep=2pt] at (100.86,15.84) {25};
\node[anchor=north,text=costmuted,inner sep=2pt] at (135.581,15.84) {50};
\node[anchor=north,text=costmuted,inner sep=2pt] at (170.3,15.84) {100};
\node[anchor=north,text=costmuted,inner sep=2pt] at (205.02,15.84) {200};
\draw[referencepurple,line width=1.2pt,dashed] (100.86,47.3)--(135.581,45.32)--(170.3,56.5387)--(205.02,59.8374);
\filldraw[fill=white,draw=referencepurple,line width=1pt] (100.86,47.3) circle (2pt);
\filldraw[fill=white,draw=referencepurple,line width=1pt] (135.581,45.32) circle (2pt);
\filldraw[fill=white,draw=referencepurple,line width=1pt] (170.3,56.5387) circle (2pt);
\filldraw[fill=white,draw=referencepurple,line width=1pt] (205.02,59.8374) circle (2pt);
\node[anchor=south,text=referencepurple] at (205.02,62.4774) {53.3};
\draw[costrobot,line width=1.2pt] (29.376,43.34)--(64.0958,61.16)--(100.86,61.8213)--(135.581,65.12);
\filldraw[fill=costrobot,draw=white,line width=.4pt] (28.152,41.58) rectangle (30.6,45.1);
\filldraw[fill=costrobot,draw=white,line width=.4pt] (62.8718,59.4) rectangle (65.3198,62.92);
\filldraw[fill=costrobot,draw=white,line width=.4pt] (99.6363,60.0613) rectangle (102.084,63.5813);
\filldraw[fill=costrobot,draw=white,line width=.4pt] (134.357,63.36) rectangle (136.804,66.88);
\node[anchor=south,text=costrobot] at (135.581,67.76) {60.0};
\node at (119.7,-3.52) {Demos per task (log scale)};

\node[rotate=90] at (236.25,49.28) {Collection labor (s)};
\draw[costgrid,line width=.35pt] (263.25,17.6)--(351.75,17.6);\node[anchor=east,text=costmuted] at (261,17.6) {0};
\draw[costgrid,line width=.35pt] (263.25,38.72)--(351.75,38.72);\node[anchor=east,text=costmuted] at (261,38.72) {50};
\draw[costgrid,line width=.35pt] (263.25,59.84)--(351.75,59.84);\node[anchor=east,text=costmuted] at (261,59.84) {100};
\draw[costgrid,line width=.35pt] (263.25,80.96)--(351.75,80.96);\node[anchor=east,text=costmuted] at (261,80.96) {150};
\draw[costmuted,line width=.4pt] (263.25,80.96)--(263.25,17.6)--(351.75,17.6);
\fill[referencepurple,rounded corners=.7pt] (275.25,17.6) rectangle (294.75,31.6659);\node[anchor=south,inner sep=2pt] at (285,32.5459) {33.30};\node[anchor=north,inner sep=2pt] at (285,15.84) {Human};
\fill[costrobot,rounded corners=.7pt] (319.5,17.6) rectangle (339,77.095);\node[anchor=south,inner sep=2pt] at (329.25,77.975) {140.85};\node[anchor=north,inner sep=2pt] at (329.25,15.84) {Robot};
\end{tikzpicture}
\endgroup}
  \caption{\textbf{Task performance and collection labor with human or robot supervision.}
  Equal-weight means over Close Laptop and Pull Curtain. Left: task scores versus demonstrations per task, evaluated over 20 trials per policy. Right: person-seconds per demonstration, including capture, amortized setup, calibration, troubleshooting, and resets, with one person for human capture and two for teleoperation.}
  \label{fig:collection-efficiency}
\end{figure}
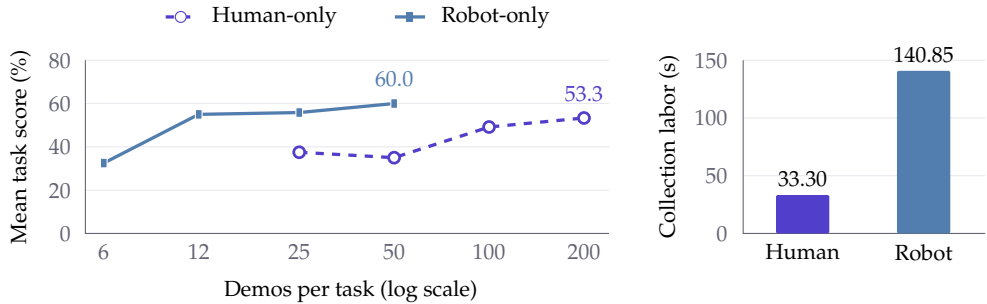

\paragraph{Effectiveness across task families.}
Figure~\ref{fig:zeroshot-alignment} covers four whole-body tasks and three long-range tasks, with mean progress of 51.7\% and 76.2\%, respectively (62.2\% when all seven tasks are weighted equally). Paired whole-body comparisons show the benefit of alignment, while long-range evaluations cover relocation, navigation, and foot interaction at seen and unseen target locations. These evaluations test policies learned from task-specific human demonstrations without target-task robot data. The following experiments test whether human-data mid-training improves \modelname{} after robot post-training.

\paragraph{Performance relative to collection effort.}
Figure~\ref{fig:collection-efficiency} shows competitive robot performance with substantially lower human-data collection costs. Averaging Close Laptop and Pull Curtain, human collection reduces mean labor per demonstration from 140.85 to 33.30 person-seconds (76.4\% less labor, or $4.23\times$ as many demonstrations per person-hour); policies trained on 200 human demonstrations reach a mean task score of 53.3\%, compared with 60.0\% from 50 robot demonstrations. For long-range object relocation, our human-data collection setup reduces on-site recording-and-reset time from 130 to 25 seconds per demonstration, yielding $5.2\times$ throughput. These measurements use different accounting conventions. For the approximate teaser summary, we use rounded time proxies of 70/33 seconds for whole-body robot/human capture and 130/25 seconds for long-range capture. Equal weighting of the two families gives about 100/29 seconds per demonstration, or $3.4\times$ throughput. The 70-second proxy approximates half the reported two-operator labor and is not a measured elapsed time; this summary is an estimate for representative tasks, not a corpus-wide measurement.

\subsection{Performance on Simulation and Real-World Tasks}
\label{sec:exp-performance}

We evaluate whether the human-supervised training recipe improves humanoid loco-manipulation across simulation and real-world tasks by comparing \modelname{} with established policy baselines.

\begin{figure}[!ht]
  \centering
  \input{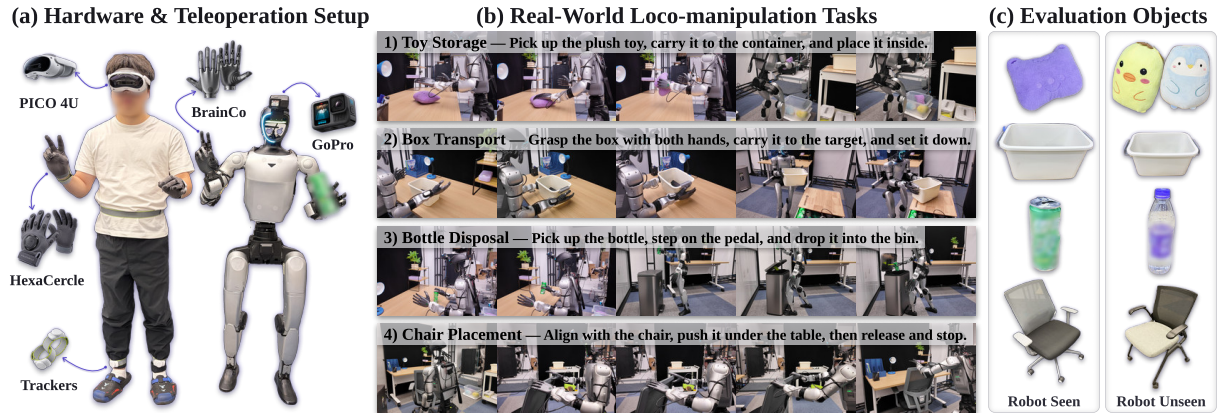}
  \setlength{\abovecaptionskip}{2pt}
  \caption{\textbf{Real-world data collection and evaluation.}
  (a) Egocentric teleoperation collects whole-body demonstrations.
  (b) Four loco-manipulation tasks evaluate task success.
  (c) Robot-seen and robot-unseen objects are used for standard task and generalization evaluation, respectively.}
  \label{fig:robot-embodiment-teleoperation}
  \label{fig:real-world-task-settings}
\end{figure}

\noindent\textbf{Hardware.}
A PICO 4 Ultra with five trackers captures body motion, which SONIC converts into Unitree G1 control signals (Fig.~\ref{fig:real-world-task-settings}(a)).
We use HexaCercle M11 gloves to capture hand poses robustly under occlusion and map the captured finger motion to control the BrainCo Revo2 hands.
A head-mounted GoPro provides egocentric RGB, synchronized with whole-body demonstrations on G1's computer.

\noindent\textbf{Evaluation.}
We evaluate six core SIMPLE tasks~\citep{wei2026simple} at Levels~0--2 with ten rollouts per task--level pair, and four real-world tasks shown in Fig.~\ref{fig:real-world-task-settings}(b), with ten frozen initial configurations per task and policy.
Fixed initial conditions support consistent policy comparisons.
We report task success and progress to capture full and partial completion.
Progress sums fixed milestone scores, normalized to 100 per task and averaged equally across tasks (Appendix~\ref{sec:app-real-tasks}).
Standard trials use robot-seen objects in Fig.~\ref{fig:real-world-task-settings}(c).
Generalization uses objects seen only in human demonstrations.

\noindent\textbf{Baselines.}
Simulation baselines come from the public SIMPLE leaderboard~\citep{wei2026simple}.
Public entries retain their source protocols. We evaluate StarVLA and \modelname{} ourselves.
Real-world baselines include $\pi_{0.5}$~\citep{physicalintelligence2025pi05}, GR00T N1.6~\citep{gear2025grootn16}, $\Psi_0$~\citep{wei2026psi0}, and StarVLA~\citep{starvlacommunity2026starvla}.
Real-world methods share initial configurations and scoring criteria; baseline implementations are detailed in Appendix~\ref{sec:app-baseline-implementation}.

\paragraph{Real-world evaluation.}

Table~\ref{tab:real-robot-results} reports the highest average success rate (62.5\%) and weighted progress (80.4\%) for \modelname{}, exceeding the strongest external baselines by 22.5\% and 25.8\%, respectively.
It achieves the highest success count on all four tasks, with the largest progress gain on bottle disposal (51.0\%), and keeps progress above 75\% on every task.
Chair placement separates the two metrics: \modelname{} completes 6/10 trials to StarVLA's 3/10 yet trails on progress (77.0\% versus 82.5\%).

\begingroup
\setlength{\intextsep}{2pt}
\setlength{\textfloatsep}{6pt}
\begin{table}[!h]
  \centering
  \caption{\textbf{Real-world task performance.}
  We report success rates over ten trials and weighted progress (\%) for each task.
  Averages weight the four tasks equally.}
  \label{tab:real-robot-results}
  \fontsize{7.7}{9.2}\selectfont
  \setlength{\tabcolsep}{1pt}
  \renewcommand{\arraystretch}{1.15}
    \begin{tabular*}{\linewidth}{@{\extracolsep{\fill}}l
      *{3}{c@{\hspace{1pt}}c}!{\hspace{3pt}}
      *{2}{c@{\hspace{1pt}}c}@{}}
      \toprule
      {\scriptsize\textbf{Method}}
        & \multicolumn{2}{c}{\makebox[0pt]{\scriptsize\textbf{Bottle Disposal}}}
        & \multicolumn{2}{c}{\makebox[0pt]{\scriptsize\textbf{Toy Storage}}}
        & \multicolumn{2}{c}{\makebox[0pt]{\scriptsize\textbf{Box Transport}}}
        & \multicolumn{2}{c}{\makebox[0pt]{\scriptsize\textbf{Chair Placement}}}
        & \multicolumn{2}{c}{\makebox[0pt]{\scriptsize\textbf{Average}}} \\
      \midrule
      $\pi_{0.5}$~\citep{physicalintelligence2025pi05}
        & 2/10 & 28.0 & 5/10 & 64.5 & 5/10 & 58.0
        & 4/10 & 65.5 & 16/40 & 54.0 \\
      GR00T N1.6~\citep{gear2025grootn16}
        & 0/10 & 10.0 & 1/10 & 19.0 & 0/10 & 10.0
        & 2/10 & 46.0 & 3/40 & 21.3 \\
      $\Psi_0$~\citep{wei2026psi0}
        & 1/10 & 26.5 & 2/10 & 30.0 & 1/10 & 37.5
        & 0/10 & 33.0 & 4/40 & 31.8 \\
      StarVLA~\citep{starvlacommunity2026starvla}
        & 0/10 & 10.0 & 5/10 & 55.0 & 3/10 & 71.0
        & 3/10 & \textbf{82.5} & 11/40 & 54.6 \\
      \midrule
      $\lambda_0$ \textbf{(Ours)}
        & \textbf{5/10} & \textbf{79.0} & \textbf{7/10} & \textbf{81.0}
        & \textbf{7/10} & \textbf{84.5} & \textbf{6/10} & 77.0
        & \textbf{25/40} & \textbf{80.4} \\
      \bottomrule
    \end{tabular*}
\end{table}

\paragraph{Simulation evaluation.}
\begin{table}
  \centering
  \caption{\textbf{SIMPLE task performance.}
  Success rates (\%) at Levels 0--2 with ten trials per task--level pair.
  Average is computed per level.
  Overall averages all 18 task--level pairs.}
  \label{tab:simple-summary}
  \setlength{\tabcolsep}{3pt}
  \resizebox{\linewidth}{!}{%
    \begin{tabular}{lcccccccc}
      \toprule
      Model & XMovePick & BendPick
      & Handover & Mobile P\&P
      & Grasp & XMoveBendPick
      & Average & Overall \\
      \midrule
      H-RDT~\citep{bi2025hrdt} & 0/0/20 & 0/0/10 & 0/10/0 & 0/0/0 & 0/0/0 & 0/0/0 & 0.0/1.7/5.0 & 2.2 \\
      InternVLA-M1~\citep{chen2025internvlam1} & 0/0/0 & 50/50/0 & 0/0/0 & 0/0/0 & 0/0/0 & 30/50/70 & 13.3/16.7/11.7 & 13.9 \\
      EgoVLA~\citep{yang2025egovla} & 0/10/20 & 70/50/80 & 0/40/30 & 0/0/0 & 100/100/70 & 30/50/40 & 33.3/41.7/40.0 & 38.3 \\
      Diffusion Policy~\citep{chi2025diffusion} & 30/30/20 & 100/80/60 & 30/20/40 & 40/0/0 & 80/90/80 & 0/0/0 & 46.7/36.7/33.3 & 38.9 \\
      GR00T N1.6~\citep{gear2025grootn16} & 100/100/70 & 70/70/60 & 10/30/30 & 0/0/0 & 90/90/70 & 40/40/10 & 51.7/55.0/40.0 & 48.9 \\
      $\pi_{0.5}$~\citep{physicalintelligence2025pi05} & 70/50/10 & 100/100/80 & 50/40/50 & 30/30/30 & 100/100/80 & 0/0/0 & 58.3/53.3/41.7 & 51.1 \\
      DreamZero~\citep{ye2026dreamzero} & 100/100/100 & 90/90/80 & 70/80/90 & 50/30/30 & 90/100/70 & 0/0/10 & 66.7/66.7/63.3 & 65.6 \\
      ACT~\citep{zhao2023aloha} & 100/100/50 & 100/90/90 & 70/70/100 & 50/50/50 & 100/100/80 & 90/100/100 & 85.0/85.0/78.3 & 82.8 \\
      $\Psi_0$~\citep{wei2026psi0} & 100/100/60 & 100/100/100 & 70/70/100 & 70/50/60 & 100/100/80 & 100/90/90 & \textbf{90.0}/85.0/81.7 & 85.6 \\
      StarVLA~\citep{starvlacommunity2026starvla} & 100/90/90 & 70/90/70 & 80/80/100 & 90/80/80 & 90/100/70 & 70/100/100 & 83.3/90.0/85.0 & 86.1 \\
      \midrule
      \rowcolor{black!4}
      \textbf{\modelname{} (Ours)} & 100/100/80 & 100/100/100 & 90/100/100 & 90/70/70 & 100/90/80 & 60/100/90 & \textbf{90.0}/\textbf{93.3}/\textbf{86.7} & \textbf{90.0} \\
      \bottomrule
    \end{tabular}%
  }
\end{table}

Table~\ref{tab:simple-summary} reports 90.0\% overall success for \modelname{}, exceeding StarVLA and $\Psi_0$ by 3.9\% and 4.4\%, respectively. This summary uses task-wise configuration selection, with limitations detailed in Appendix~\ref{sec:app-simple}.
It leads StarVLA at every level (90.0\%, 93.3\%, and 86.7\% at Levels~0--2), though the margin narrows to 1.7\% at Level~2.
It reaches 100\% on BendPick at all levels and on Handover at Levels~1 and 2; Mobile P\&P is the weakest case, trailing StarVLA at Levels~1 and 2 (70\% versus 80\%).

\vspace{-0.8ex}
\subsection{Human-Data Scale and Training-Stage Ablations}
\label{sec:exp-learning}

We vary Stage-II data to measure validation loss and downstream task progress, then ablate the two human-data stages under fixed robot supervision.

\paragraph{Validation-loss scaling.}

To isolate human-data scale in Stage II, we train the same 2B model for 100k updates on nested 5\%, 10\%, 25\%, 50\%, and 100\% subsets with a fixed seed and optimization schedule (Appendix~\ref{sec:app-scaling}).
Every 1k updates, we evaluate two fixed, session-disjoint holdouts: \emph{clean} contains sessions with upper-quartile whole-body and bilateral-hand annotation validity, while \emph{representative} is stratified by data source and annotation validity to reflect the full data distribution.
Following EgoScale~\citep{zheng2026egoscale}, we analyze training dynamics and minimum loss at each scale.

\setlength{\intextsep}{4pt}
\begin{figure}[!htbp]
  \centering
  \includegraphics[width=\linewidth,trim=0 7bp 0 13bp,clip]{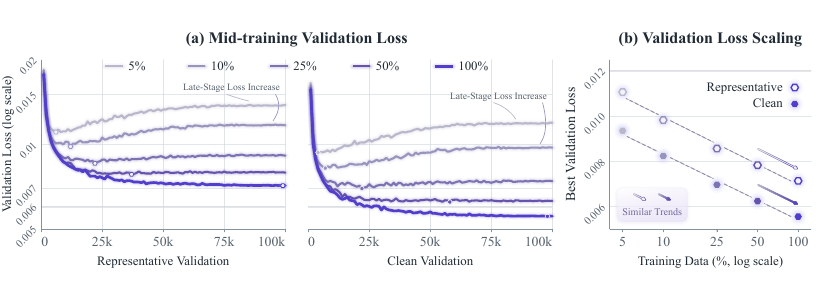}\par
  \setlength{\abovecaptionskip}{2pt}
  \caption{\textbf{Human-data scaling of validation loss.}
  (a) Training curves on representative and clean holdouts at five data fractions.
  (b) Minimum validation loss versus data scale with descriptive log-linear fits over the measured range.
  Minima are selected independently on each holdout.}
  \label{fig:scaling-curves}
\end{figure}
\setlength{\intextsep}{2pt}

\vspace{-6pt}
\paragraph{Training dynamics.}
Fig.~\ref{fig:scaling-curves}(a) shows later validation-loss minima at larger data scales.
At 5\% and 10\%, representative loss reaches its minimum at 6k and 12k updates, then finishes 24.7\% and 19.0\% higher.
On clean, the increases are 27.6\% and 18.1\%.
At 50\% and 100\%, final losses remain within 1.4\% of the minima on representative and 0.9\% on clean.
At 50\%, the minima occur at 37k/58k updates on representative/clean. At 100\%, they occur at 99k/98k.
The 25\% run forms an intermediate regime: both holdouts reach their minima near 22k updates and finish only 6.4--6.5\% higher.
Larger datasets thus support longer training before validation loss rises, while smaller subsets benefit from earlier checkpoint selection.
Final checkpoints alone conflate data-scale effects with late-stage loss increases.

\vspace{-6pt}
\paragraph{Scaling trend.}
Fig.~\ref{fig:scaling-curves}(b) compares minimum validation losses across data scales.
From 5\% to 100\%, loss falls by 35.5\% on representative and 40.6\% on clean.
Fitting $\widehat{L}_h(D)=a_h+b_h\ln D$ gives $R^2=0.9868/0.9901$ and per-doubling reductions of $0.000900/0.000877$ for representative/clean.
Clean has lower loss throughout, while slope magnitudes differ by only 2.7\%, indicating similar scaling responses despite their different annotation-validity distributions.
These single-run fits describe the measured range without establishing cross-seed reproducibility or extrapolation.

\noindent\begin{minipage}[t]{0.53\linewidth}
\vspace{0pt}
\noindent\textbf{Downstream task-progress scaling.}
Fig.~\ref{fig:downstream-scaling} reports task progress of 59.9\%, 64.2\%, 67.3\%, 72.1\%, 76.4\%, and 80.4\% at 0\%, 5\%, 10\%, 25\%, 50\%, and 100\% Stage-II data, respectively.
The endpoints match the no-Stage-II and full-recipe results in Table~\ref{tab:real-robot-ablation}.
We fix Stage-I initialization, model capacity, robot demonstrations, Stage-III update budget, and evaluation configurations.
Nonzero fractions share Stage-II update budgets and checkpoint selection; Appendix~\ref{sec:app-scaling} details subset construction and fixed controls.
Progress increases monotonically across these fractions, gaining 20.5\% from no Stage II to the full corpus.
\end{minipage}\hfill
\begin{minipage}[t]{0.43\linewidth}
\vspace{0pt}
\begingroup
  \definecolor{downstreamink}{HTML}{202A34}
  \definecolor{downstreammuted}{HTML}{55616C}
  \definecolor{downstreamgrid}{HTML}{E2E7EC}
  \setlength{\parskip}{0pt}
  \centering
  \begin{tikzpicture}
    \begin{axis}[
      width=\linewidth,
      height=4.1cm,
      xmin=0, xmax=5,
      ymin=59, ymax=90,
      xtick={0,1,2,3,4,5},
      xticklabels={0\%,5\%,10\%,25\%,50\%,100\%},
      ytick={60,70,80,90},
      xlabel={Stage-II data fraction},
      ylabel={Task progress (\%)},
      tick label style={font=\scriptsize,text=downstreammuted},
      label style={font=\scriptsize,text=downstreamink},
      axis x line*=bottom,
      axis y line*=left,
      axis line style={downstreammuted!75},
      tick style={downstreammuted!75},
      tick align=outside,
      grid=major,
      grid style={downstreamgrid,line width=.35pt},
      clip=false,
    ]
      \addplot[color=referencepurple, line width=4pt, opacity=.045,
        no marks, line cap=round, line join=round]
        coordinates {(0,59.9) (1,64.2) (2,67.3) (3,72.1) (4,76.4) (5,80.4)};
      \addplot[color=referencepurple, line width=2.5pt, opacity=.085,
        no marks, line cap=round, line join=round]
        coordinates {(0,59.9) (1,64.2) (2,67.3) (3,72.1) (4,76.4) (5,80.4)};
      \addplot[color=referencepurple, line width=1.15pt,
        no marks, line cap=round, line join=round]
        coordinates {(0,59.9) (1,64.2) (2,67.3) (3,72.1) (4,76.4) (5,80.4)};
      \addplot[only marks, mark=*, mark size=5pt, draw=none,
        fill=referencepurple, opacity=.05]
        coordinates {(0,59.9) (1,64.2) (2,67.3) (3,72.1) (4,76.4) (5,80.4)};
      \addplot[only marks, mark=*, mark size=3.5pt, draw=none,
        fill=referencepurple, opacity=.11]
        coordinates {(0,59.9) (1,64.2) (2,67.3) (3,72.1) (4,76.4) (5,80.4)};
      \addplot[only marks, mark=*, mark size=2pt,
        draw=white, line width=.4pt, fill=referencepurple]
        coordinates {(0,59.9) (1,64.2) (2,67.3) (3,72.1) (4,76.4) (5,80.4)};
      \node[font=\scriptsize, text=referencepurple, anchor=south west]
        at (axis cs:0,59.9) {59.9};
      \node[font=\scriptsize, text=referencepurple, anchor=south, yshift=2pt]
        at (axis cs:1,64.2) {64.2};
      \node[font=\scriptsize, text=referencepurple, anchor=south, yshift=2pt]
        at (axis cs:2,67.3) {67.3};
      \node[font=\scriptsize, text=referencepurple, anchor=south, yshift=2pt]
        at (axis cs:3,72.1) {72.1};
      \node[font=\scriptsize, text=referencepurple, anchor=south, yshift=2pt]
        at (axis cs:4,76.4) {76.4};
      \node[font=\scriptsize, text=referencepurple, anchor=south east]
        at (axis cs:5,80.4) {80.4};
    \end{axis}
  \end{tikzpicture}\par
  \vspace{-4pt}
  \setlength{\abovecaptionskip}{0pt}
  \captionof{figure}{Average real-world task progress versus Stage-II human-data fraction.}
  \label{fig:downstream-scaling}
\endgroup

\end{minipage}\par

\vspace{-6pt}
\paragraph{Training-stage ablations under fixed robot supervision.}
Table~\ref{tab:real-robot-ablation} ablates the two human-data stages while retaining the same architecture, action representations, vision--language initialization, and Stage-III robot post-training in every variant. SIMPLE success is 90.0\% with the full recipe, 87.8\% without Stage I, 82.3\% without Stage II, and 79.4\% without both.
Real-world success/progress is 62.5\%/80.4\% for the full recipe, 50.0\%/75.9\% without Stage I, 22.5\%/59.9\% without Stage II, and 15.0\%/54.0\% without both.
Removing Stage II reduces real-world success far more than removing Stage I (40.0\% versus 12.5\%): whole-body mid-training contributes more than the interaction priors from egocentric hand data.

\begin{table}[!htbp]
  \centering
  \caption{\textbf{Training-stage ablations with fixed robot supervision.}
  Filled/open circles indicate included/removed data. Metrics report SIMPLE success and real-world averages across four tasks. SIMPLE configuration-selection limitations are detailed in Appendix~\ref{sec:app-simple}.}
  \label{tab:real-robot-ablation}
  \small
  \setlength{\tabcolsep}{2.2pt}
  \resizebox{0.9\linewidth}{!}{%
    \begin{tabular*}{\linewidth}{@{\extracolsep{\fill}}lccccccccc@{}}
      \toprule
      \multirow{2}{*}{Variant}
        & \multirow{2}{*}{S1}
        & \multirow{2}{*}{S2}
        & \multirow{2}{*}{Robot}
        & \multicolumn{4}{c}{SIMPLE (\%)}
        & \multicolumn{2}{c}{Real-world (\%)} \\
      \cmidrule(lr){5-8}\cmidrule(lr){9-10}
        & & & & L0 & L1 & L2 & Overall & Success & Progress \\
      \midrule
      w/o human-data pretraining
        & $\circ$ & $\circ$ & $\bullet$ & 83.3 & 83.3 & 71.7 & 79.4 & 15.0 & 54.0 \\
      w/o Stage II mid-training
        & $\bullet$ & $\circ$ & $\bullet$ & 83.6 & 86.7 & 76.7 & 82.3 & 22.5 & 59.9 \\
      w/o Stage I pretraining
        & $\circ$ & $\bullet$ & $\bullet$ & 88.0 & 91.7 & 83.7 & 87.8 & 50.0 & 75.9 \\
      $\lambda_0$ \textbf{(Full Recipe)}
        & $\bullet$ & $\bullet$ & $\bullet$ & \textbf{90.0} & \textbf{93.3} & \textbf{86.7} & \textbf{90.0} & \textbf{62.5} & \textbf{80.4} \\
      \bottomrule
    \end{tabular*}%
  }
\end{table}

\FloatBarrier

\endgroup
\subsection{Transfer to Robot-Unseen Object Instances}
\label{sec:exp-transfer}

To assess generalization, we evaluate whether human demonstrations improve performance on object instances absent from robot demonstrations across the four real-world tasks.
For each method, we train a policy using robot demonstrations of object instances A and human demonstrations of A and B.
\textit{Seen}~(\transferinline{1}) and \textit{Human-guided}~(\transferinline{2}) evaluate this same policy on A and B, respectively.
\textit{Robot-only}~(\transferinline{3}) evaluates a robot-only variant on B under matched initializations and scoring. Unlike Sec.~\ref{sec:exp-zeroshot}, this setting permits robot demonstrations of the same task on A, while testing transfer to B.

\begingroup
\setlength{\intextsep}{8pt}
\begin{figure}[!ht]
  \centering
  \setlength{\abovecaptionskip}{0pt}
  \makebox[\linewidth][l]{\hspace*{-0.094\linewidth}\resizebox{1.095\linewidth}{!}{\input{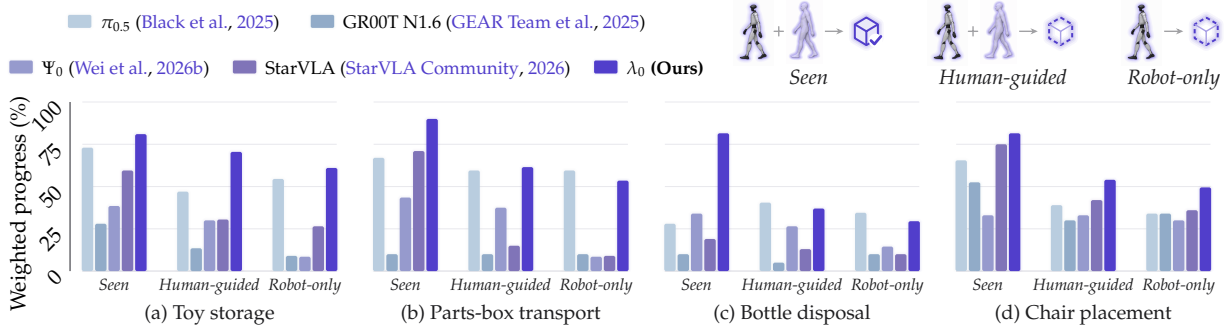}}}
  \caption{\textbf{Generalization with human demonstrations.}
  Seen and Human-guided evaluate the same policy, trained on robot demonstrations of A and human demonstrations of A and B, on A and B respectively.
  Robot-only evaluates a policy trained without human data, using robot demonstrations of A, on B. B is absent from robot demonstrations in both training recipes; it is human-observed only in the joint-data recipe.
  Solid checked and dashed object icons indicate presence and absence in robot demonstrations, respectively.
  Seen scores use separate trials from Table~\ref{tab:real-robot-results}.}
  \label{fig:generalization-results}
\end{figure}

\endgroup

\modelname achieves the highest mean \textit{Human-guided} progress at 55.8\%, compared with 46.5\% for the strongest baseline, $\pi_{0.5}$ (Fig.~\ref{fig:generalization-results}).
Compared with its \textit{Robot-only} variant, \modelname improves progress on all four tasks, increasing the average from 48.4\% to 55.8\%, a gain of 7.4\%.
Progress on B remains below the \textit{Seen} setting. The comparison measures the combined benefit of human-data training; it does not isolate the contribution of human B demonstrations from human A demonstrations or earlier human pretraining, nor does it vary human-data scale.

\section{Conclusion}
We presented \modelname, a foundation model for humanoid loco-manipulation, and \datasetname, an egocentric whole-body human dataset that supports learning from open-world interaction without robot-operated collection.
Human-to-humanoid alignment makes this experience usable for learning coordinated body and hand actions.
Whole-body human mid-training bridges diverse human interaction videos and limited, costly robot demonstrations, progressively extending hand-motion pretraining to whole-body control before adaptation to the target humanoid.
Our results show that scaling this human supervision improves downstream loco-manipulation, supporting egocentric whole-body human experience as a practical source of data for humanoid learning.
%

\section{Contributions}
\label{sec:contributions}
Within each category, authors are listed \textbf{alphabetically} by first name.

\paragraph{Core Contributors.}
Chongyang Xu, Jin Chen, Jinhui Ye, Yilun Chen{\def\thefootnote{\dag}\footnote{Project Lead.}}, Yiming Jiang, Zhao Wu

\paragraph{Contributors.}
Hu Wei, Lin Qu, Shifeng Zhang, Yisheng He, Yuming Jiang, Zenan Meng, Zheng Wu

\paragraph{Supervisor.}
Mu Xu, Steven Hoi

\paragraph{Acknowledgments.}
We gratefully acknowledge the Data Technology Team at Alibaba Inc. and the Embodied AI Data Team at Amap, Alibaba Inc., for their support with data collection. We are especially grateful to Anjie Ma, Jiancheng Zhu, Junjie Liu, Lan Yang, Lei Pan, Suyu Wang, Yongtuo Zhao, Zenan Meng, and Zheng Wu. We also thank the following data collectors for their work in collecting the data: Fangzheng Diao, He Zhao, Hongbo Pan, Jiale Wang, Niannian Zhang, Ruiwei Zhang, Shaofei Song, Shengquan Jiang, Shitai Gan, Siyu Zhu, Xinhao Sun, Yafei Wu, Yi Shi, Yu Xu, Zehong Li, and Zejun Ren.

\begingroup
\hypersetup{urlcolor=black,linkcolor=black,citecolor=black}
\let\originalbibsection\bibsection
\renewcommand{\bibsection}{%
  \originalbibsection
  \phantomsection
  \addcontentsline{toc}{section}{References}%
}
\bibliography{references}
\bibliographystyle{styles/tongyi_report}
\endgroup

\clearpage
\appendix
\section*{Appendix}
\phantomsection
\addcontentsline{toc}{section}{Appendix}
We provide a project website and an overview video to accompany the paper.
The supplementary material below provides data-quality and split-validation details, additional training formalism and implementation details, baseline implementations, and validation-scaling analysis details.

\section{Related-Work Comparisons}
\label{sec:app-data}

Tables~\ref{tab:egocentric-datasets} and~\ref{tab:closest-work} compare datasets and methods in capture, motion supervision, training, and scaling.
\begin{table}[!htbp]
  \centering
  \caption{Egocentric human-data resources and documented training interfaces.
  Body and hand coverage denotes motion supervision.
  NR denotes an unreported interface, not an incompatible one.}
  \label{tab:egocentric-datasets}
  \begingroup
  \fontsize{8}{9.5}\selectfont
  \setlength{\tabcolsep}{2pt}
  \renewcommand{\arraystretch}{1.13}
  \begin{tabular}{@{}>{\raggedright\arraybackslash}p{75pt}
    >{\raggedright\arraybackslash}p{30pt}
    >{\raggedright\arraybackslash}p{66pt}
    >{\raggedright\arraybackslash}p{44pt}
    >{\raggedright\arraybackslash}p{73pt}
    >{\raggedright\arraybackslash}p{88pt}@{}}
    \toprule
    Resource & Hours & Body / hands & SMPL & Humanoid interface & Capture setup \\
    \midrule
    EgoDex\newline\citep{hoque2025egodex}
      & 829 & Upper body and articulated hands & NR & NR in dataset release & Apple Vision Pro \\
    \addlinespace[3pt]
    HOT3D\newline\citep{banerjee2025hot3d}
      & 13.9 & Hands and objects without full-body coverage & NR & NR & Aria / Quest 3 with external tracking \\
    \addlinespace[3pt]
    NymeriaPlus\newline\citep{detone2026nymeriaplus}
      & NR$^{a}$ & Full body, finger motion NR & Provided & NR & Aria, wristbands, XSens \\
    \addlinespace[3pt]
    EgoHumanoid\newline\citep{shi2026egohumanoid}
      & NR & Full body and articulated hands & NR & G1: wrist targets + locomotion commands & PICO + trackers, ZED X Mini \\
    \addlinespace[3pt]
    EgoScale\newline\citep{zheng2026egoscale}
      & 20,854 & Wrist and hand action supervision & NR & G1 transfer via wrist / hand interface & Multiple human-data sources \\
    \addlinespace[3pt]
    Index\newline\citep{figure2026index}
      & NR & Pose schema NR & NR & Figure / Helix, G1 NR & Distributed creator recordings \\
    \addlinespace[3pt]
    EgoSuite-Open100K\newline\citep{lightwheel2026egosuite}
      & 10k$^{b}$ & Hands, full body in body subsets & NR & NR (LeRobot is a storage format) & Head camera, wrist views in Pro \\
    \midrule
    \rowcolor{black!4}
    \textbf{\datasetname}\newline\textbf{(Ours)}
      & 500 & Tracked full body and reconstructed hands & Compatible & SONIC body tokens and G1 training views & Portable PICO + GoPro \\
    \bottomrule
  \end{tabular}
  \par\vspace{3pt}
  \begin{minipage}{\linewidth}
    \fontsize{7.5}{9}\selectfont $^{a}$NymeriaPlus upgrades the 300+ h Nymeria corpus. Its retained duration after sequence filtering is not separately specified here.
    $^{b}$The August 26, 2026 announcement reports 10k h initially available out of 100k h planned, including 12k h of planned full-body subsets. These planned hours are not counted as released full-body data.
    SMPL compatibility and a demonstrated robot interface are distinct.
    Our body stream is device-derived kinematics, not optical motion-capture ground truth. Hand geometry is reconstructed offline.
    Capture hardware is listed without an unverified cost ranking.
  \end{minipage}
  \endgroup
\end{table}

\FloatBarrier

\begin{table}[!h]
  \centering
  \caption{Problem-level comparison with the closest humanoid VLA and world-action studies.
  ``General'' denotes human data not collected for the downstream evaluation tasks. ``Task-targeted'' denotes demonstrations whose task or object coverage is selected for the downstream study.
  This table compares experimental questions rather than reported scores.}
  \label{tab:closest-work}
  \scriptsize
  \setlength{\tabcolsep}{3.2pt}
  \resizebox{\linewidth}{!}{%
  \begin{tabular}{@{}p{2.05cm}p{3.45cm}p{3.35cm}p{3.25cm}p{3.65cm}@{}}
    \toprule
    Method & Human signal & Role in training & Controlled scale & Primary question \\
    \midrule
    \makecell[l]{WholeBodyVLA\\\citep{jiang2026wholebodyvla}}
      & 300 h action-free egocentric video
      & latent locomotion pre-training, robot manipulation latents, and teleop grounding
      & 0--100\% video $\times$ teleop trajectories
      & Can visual latent actions reduce loco-manipulation teleoperation? \\
    \makecell[l]{EgoHumanoid\\\citep{shi2026egohumanoid}}
      & same-task RGB and wearable motion
      & direct human--robot co-training after view/action alignment
      & 0--300 human demonstrations and sampling ratio
      & Can task-matched human data improve scene generalization? \\
    \makecell[l]{$\Psi_0$\\\citep{wei2026psi0}}
      & 829 h wrist/fingertip-labeled EgoDex video
      & task-space VLM pre-training before robot joint-action learning
      & full EgoDex versus 10\%
      & Which staged data recipe yields an effective humanoid foundation model? \\
    \makecell[l]{OpenHLM\\\citep{hu2026openhlm}}
      & task-targeted stationary HuMI motion
      & heterogeneous co-training after whole-body VLA adaptation
      & whole-body teleop count and data-source choice
      & Which interfaces and data sources yield a whole-body-native VLA? \\
    \makecell[l]{$\omega$-0\\\citep{li2026omega0}}
      & public human video and whole-body motion (ARCTIC, Xperience-10M, Motion-X)
      & action-token VLM pre-training and SONIC replay for predictive action-latent learning
      & human-data volume scaling not reported; data-source and model ablations
      & Can latent visual foresight improve concurrent whole-body loco-manipulation? \\
    \textbf{\modelname}
      & 500 h synchronized RGB, full-body motion, and bilateral hands
      & human-data mid-training followed by robot post-training
      & 0--100\% Stage-II human data with a fixed robot budget
      & What scales with explicit whole-body human experience under fixed grounding? \\
    \bottomrule
  \end{tabular}%
  }
\end{table}

\FloatBarrier

\section{Action Interfaces and Training Details}
\label{sec:app-training-details}

We define Stage-I hand-state tokens and the G1-compatible supervision for Stages II and III.

\subsection{State--Action Definitions}
\label{sec:app-action-interfaces}

\paragraph{Egocentric hand-state token interface.}
For hand $\chi\in\{L,R\}$, let $\mathbf{p}_t^\chi\in\mathbb{R}^3$ and $\mathbf{R}_t^\chi\in\mathrm{SO}(3)$ denote the wrist pose, let $\rho_6(\mathbf{R})\in\mathbb{R}^6$ be its continuous 6D representation, and let $\mathbf{g}_t^\chi\in\mathbb{R}^{60}$ contain the 20 non-wrist joints in a wrist-local, palm-aligned, scale-normalized frame.
The current hand state is
\begin{equation}
  \mathbf{h}_t =
  \left[\mathbf{p}_t^L,\rho_6(\mathbf{R}_t^L),\mathbf{g}_t^L,
  \mathbf{p}_t^R,\rho_6(\mathbf{R}_t^R),\mathbf{g}_t^R\right]
  \in\mathbb{R}^{138}.
  \label{eq:hand-state}
\end{equation}
We follow the state discretization of $\pi_{0.5}$~\citep{physicalintelligence2025pi05}: each normalized scalar is assigned to one of 256 uniform bins over $[-1,1]$, and the bin indices are serialized as text and passed through the vision--language tokenizer. We apply this rule to the current hand state in the input and to each hand state in the future sequence used as the Stage-I prediction target. Stage I predicts future hand-state tokens without an action expert or continuous action head.

\paragraph{G1-compatible whole-body interface.}
Let $\mathbf{z}_t\in\mathbb{R}^{64}$ be the motion token produced by the frozen SONIC encoder.
For converted human data, a closed-loop free-base G1 rollout provides 29 body-joint positions and projected gravity $\mathbf{g}^{\mathrm{proj}}_t\in\mathbb{R}^3$, while hand retargeting provides six normalized Revo2 commands per side.
Its interface is
\begin{equation}
  \begin{aligned}
  \mathbf{s}_t^{R} & =
  \left[\mathbf{q}_t^{1:29},\mathbf{u}_t^L,\mathbf{u}_t^R,
  \mathbf{g}^{\mathrm{proj}}_t\right]\in\mathbb{R}^{44},\\
  \mathbf{a}_{t+h}^{R} & =
  \left[\mathbf{z}_{t+h},\mathbf{u}_{t+h}^L,\mathbf{u}_{t+h}^R\right]
  \in\mathbb{R}^{76}.
  \end{aligned}
  \label{eq:robot-interface}
\end{equation}
Native G1 post-training data use the same state--action format, with distinct domain identifiers and separate input and output projections.
Other embodiments, including SIMPLE, retain their native state and actuator blocks.
Variable-width states and actions are padded only at collation time. Masks remove padding, episode boundaries, failed conversions, and invisible hands from supervision.

\subsection{Model Architecture}
\label{sec:app-layerwise-policy}
\label{sec:app-denoising-details}

\paragraph{Backbone and action expert.}
The Qwen3.5-2B backbone supplies projected features to each layer of the action expert. Domain-specific input projections encode noisy actions and sinusoidal noise embeddings, while output projections predict action velocities. Following X-VLA~\citep{zheng2025xvla}, we use embodiment-conditioned soft prompts; we use 32 prompt tokens and a learned embodiment bias.

\paragraph{Action prediction.}
Stages II and III use flow matching following $\pi_{0.5}$~\citep{physicalintelligence2025pi05}. The loss averages squared velocity error over valid scalar targets, excluding missing annotations, padding, and action prefixes, without action-block reweighting. For real-time chunking (RTC), half of the training samples have no prefix; the remainder use up to eight valid prefix steps, kept noise-free and excluded from the loss. Each observation uses four independent noise realizations during mid-training and post-training. Stage I predicts discrete hand-state tokens within the backbone.

\subsection{Training Configuration}
\label{sec:app-stage-recipe}

Table~\ref{tab:app-training-config} summarizes the default 2B training schedule. EMA updates follow each optimizer step with decay $0.999$. The Stage-I backbone EMA initializes Stage II; the Stage-II policy EMA initializes Stage III.

\begin{table}[!htbp]
  \centering
  \caption{Default training schedule for the 2B model. Stages II and III predict 50-step action chunks.}
  \label{tab:app-training-config}
  \small
  \setlength{\tabcolsep}{5pt}
  \begin{tabular}{@{}lll@{}}
    \toprule
    Stage & Optimizer updates & Updated modules \\
    \midrule
    I & 200k & Vision--language backbone \\
    II & 100k & Backbone, action expert, human-data projections \\
    III & 50k & Backbone, action expert, robot interfaces and conditioning \\
    \bottomrule
  \end{tabular}
\end{table}

Stage I uses batch size 1024, learning rate $10^{-5}$, and a $0.1$ backbone learning-rate multiplier. We sample the seven Stage-I datasets in proportion to the square root of their usable frame counts. VITRA-1M additionally indexes activity from Ego4D, EPIC-KITCHENS, Ego-Exo4D, and Something-Something V2~\citep{grauman2022ego4d,damen2022epic,grauman2024egoexo4d,goyal2017something}.

\FloatBarrier

\section{Data Quality and Scaling Protocol}
\label{sec:app-scaling}
\label{sec:dataset-validation}
\label{sec:app-logged-scaling}

\paragraph{Data quality.}
A frozen episode manifest records provenance, capture metadata, annotation coverage, conversion versions, and split assignments. Structural checks verify timestamps, stream overlap and lengths, coordinate conventions, and finite valid values. Hand reprojection into RGB and synchronized visual inspection assess geometric consistency. Invalid targets are masked, and reconstruction diagnostics and conversion failures are retained. Dataset summaries cover corpus size and diversity, annotation coverage, and conversion failures; visual checks provide qualitative reconstruction diagnostics.

\paragraph{Data splits and validation.}
We partition \datasetname by recording session before extracting temporal windows, so all windows from an episode inherit the same split. Task-specific human demonstrations collected for downstream post-training remain separate from the general-purpose mid-training corpus. Session, object-instance, and evaluation-configuration identifiers track experimental separation; human-data selection does not consult the robot evaluation manifest.

\emph{Clean} contains held-out sessions whose frame-weighted validity ratios for both SONIC whole-body and bilateral hand annotations reach their respective upper-quartile thresholds in the original split. Here, clean refers to annotation validity rather than appearance or task difficulty. \emph{Representative} is stratified by data source and quartile bins of the two validity ratios.
The sets contain 1,008 and 513 episodes, respectively, each from 44 capture sessions. Their sessions are mutually disjoint and excluded from training; both sets remain fixed across data fractions.

\paragraph{Scaling protocol.}
The human-data scaling axis is expressed as a fraction of the frozen \datasetname training partition, with validation sessions excluded before subset construction. The 0\%, 5\%, 10\%, 25\%, 50\%, and 100\% conditions define the evaluation grid for Fig.~\ref{fig:downstream-scaling}. Nonzero subsets are nested, sampled at the episode level, and stratified by activity and recording session before chunk extraction. The Stage-I checkpoint, robot post-training set, Stage-III update budget, validation sets, and evaluation manifest remain fixed. The primary compute-matched comparison uses the same number of Stage-II optimizer updates for every nonzero fraction; the 0\% reference omits Stage II.

Fig.~\ref{fig:scaling-curves} reports the 2B model at the five nonzero fractions, each trained for 100k updates with seed 43. Total validation loss is evaluated every 1k updates with validation seed 20260813.

Following EgoScale~\citep{zheng2026egoscale}, we analyze validation dynamics and minimum loss separately. Each holdout's minimum is selected independently over the full 0--100k-update window, so the selected steps may differ. We fit $\widehat{L}(D)=a+b\ln D$ by unweighted ordinary least squares to the five measured minima, where $D\in\{5,10,25,50,100\}$ is the training-data percentage.
Each fraction has one run, without across-seed error bars. Fits describe only the observed range; their $R^2$ values measure in-sample fit, without establishing cross-seed reproducibility, equal slopes, or extrapolation accuracy. Similar absolute improvements need not imply similar relative improvements. Closed-loop robot performance is evaluated separately from these offline prediction metrics.

\section{Evaluation Protocols}

\subsection{Baseline Implementations}
\label{sec:app-baseline-implementation}
We adapt the four baselines to the G1 interface in Eq.~\ref{eq:robot-interface}, using absolute actions sampled at 50~Hz and a 50-frame future buffer. Dataset-specific normalization uses training statistics. Each baseline retains its upstream training implementation and checkpoint format.

\textbf{GR00T N1.6.}  We use the official Eagle vision--language backbone and N1.6 flow action expert with its 128-dimensional action-expert capacity.  The action projector is reinitialized when its dimensions change. Training uses a 50-step action horizon, learning rate $10^{-4}$, and effective batch size 64.  We initialize from the official pretrained GR00T N1.6 weights.

\textbf{$\Psi_0$.}  We use Qwen3-VL-2B, the $\Psi_0$ flow expert, and SonicTrainer with a 30-frame SONIC action chunk and RTC enabled with maximum delay 8.  We freeze the VLM and adapt the action expert with learning rate $10^{-4}$, 1{,}000 warm-up steps, cosine decay, and released backbone and action-expert checkpoints.

\textbf{$\pi_{0.5}$.}  We use the openpi JAX implementation with the official pretrained $\pi_{0.5}$ parameters, an effective batch size of 32, and the native action-sampling code.

\textbf{StarVLA.}  We use Qwen3-VL-4B with the QwenPI-v3 layerwise flow-matching action model, a 50-step action horizon, four noise realizations per observation, and ten inference steps.  The released configuration uses learning rates $10^{-5}$ for the VLM and $10^{-4}$ for the action model, cosine decay, gradient clipping at 1, and a 50k-step budget. The action expert is randomly initialized unless a checkpoint is supplied.

\subsection{Simulation Evaluation}
\label{sec:app-simple}

We evaluate the six SIMPLE tasks in Table~\ref{tab:simple-summary}, with ten rollouts per task at each of Levels 0--2. Success rates are averaged equally across tasks and levels. \modelname{} uses single-task training. StarVLA and \modelname{} results come from our evaluations; public baselines retain their source protocols.

The \modelname{} summary selects the best recorded configuration per task, without a shared checkpoint or inference setting.
All six SIMPLE tasks use the velocity-field action head.
The six tasks yield 162 successes in 180 trials.
Configuration selection was not established here as validation-only. These numbers should not be interpreted as a fixed-configuration comparison.
The Full Recipe row in Table~\ref{tab:real-robot-ablation} uses the same task-wise summary. Matched checkpoint and inference selection across ablations is unverified.

Public baseline counts follow the official leaderboard accessed September 13, 2026, including its updated ACT results.
Mobile P\&P corresponds to locomotion pick between tables.

\subsection{Real-World Evaluation and Scoring}
\label{sec:app-real-tasks}

\paragraph{Evaluation setup.}
\label{sec:app-statistics}
Each policy is evaluated on ten fixed initial configurations per task, with shared objects, target regions, instructions, timeouts, and success criteria. Configuration order is randomized per policy; trial videos and state logs retain identifiers, while manual adjudicators are blinded to model identity. We report success counts and mean progress over the ten trials, averaging equally across tasks without pooling training seeds or reporting seed-based confidence intervals. This protocol applies to the baseline comparisons and training-stage ablations in Tables~\ref{tab:real-robot-results} and~\ref{tab:real-robot-ablation}.

\paragraph{Progress and autonomous success.}
\label{sec:app-weighted-progress}
Progress sums the points for completed stages in Table~\ref{tab:real-robot-protocol}, totaling 100 per task; each stage is scored independently of preceding stages, with the same weights across methods and data scales. Contact alone does not count as a stable grasp, and lifting requires clearance from the supporting surface. For bottle disposal, approach requires the hand to enter the grasp region, arrival requires a pose suitable for pedal operation and disposal, and pedal contact scores only if the lid opens enough to admit the bottle. The five-point stability stage requires three seconds of unsupported standing without unintended collision. Full-task success is recorded separately and requires the table's terminal condition to hold for three seconds. Human contact with the robot, object, or environment after execution begins disqualifies autonomous success; assisted continuations are reported separately, and safety and intervention are logged for every task.

\begin{table}[!ht]
  \centering
  \caption{Operational definitions for the real-world evaluation.
  Progress is the sum of point values for completed stages (shown in parentheses, totaling 100 per task). Terminal success additionally requires the task-specific predicate to remain satisfied for three seconds.}
  \label{tab:real-robot-protocol}
  \scriptsize
  \setlength{\tabcolsep}{3.5pt}
  \begin{tabular}{p{0.16\linewidth}p{0.55\linewidth}p{0.22\linewidth}}
    \toprule
    Task & Scored task stages (points) & Terminal success predicate \\
    \midrule
    Bottle disposal
      & 1. Approach the bottle (10)\newline
        2. Grasp stably (15)\newline
        3. Lift clear of the table (15)\newline
        4. Arrive at the bin (15)\newline
        5. Press the pedal to open the lid (20)\newline
        6. Release the bottle into the bin (20)\newline
        7. Remain stable after disposal (5)
      & Bottle fully inside the trash can and robot standing without contact
        assistance \\
    Single-hand toy storage
      & 1. Approach the toy (10)\newline
        2. Establish contact (15)\newline
        3. Grasp stably (20)\newline
        4. Lift clear of the table (15)\newline
        5. Transport above the box (15)\newline
        6. Place inside the box (20)\newline
        7. Release with the toy remaining inside (5)
      & Toy fully inside the box, released, and not displaced during hand
        withdrawal \\
    Bimanual parts-box transport
      & 1. Approach the box (10)\newline
        2. Establish bilateral contact (15)\newline
        3. Establish a stable grasp (20)\newline
        4. Lift clear of the table (15)\newline
        5. Transport above the target surface (20)\newline
        6. Place on the target surface (15)\newline
        7. Release with the box remaining stable (5)
      & Box inside the target region, both hands released, and no contained
        parts spilled \\
    Chair placement
      & 1. Approach the chair (10)\newline
        2. Turn and align with the chair (15)\newline
        3. Establish bilateral contact (10)\newline
        4. Grasp stably and initiate pushing (20)\newline
        5. Push continuously toward the table (20)\newline
        6. Complete insertion under the table (20)\newline
        7. Release and stop safely (5)
      & Chair footprint, position, and yaw satisfy pre-registered under-table
        tolerances, with the robot released and stable \\
    \bottomrule
  \end{tabular}
\end{table}

\FloatBarrier

\paragraph{Generalization evaluation.}
The two training recipes share the same robot demonstrations of A. Human-guided and Robot-only evaluations use identical B instances, instructions, scenes, target regions, initial robot configurations, and success criteria. Figure~\ref{fig:generalization-results} reports mean weighted progress over ten trials per task, method, and condition across all four tasks (600 trials in total), using the scoring protocol above.

\stopcontents[overview]

\end{document}